\documentclass[sigconf,nonacm,balance=false]{acmart} 
\usepackage{popets}

\usepackage{hyperref}
\usepackage{url}
\usepackage{booktabs}
\usepackage{graphicx}
\usepackage{multirow}
\usepackage{makecell}
\usepackage{subcaption}

\usepackage{algorithm}
\usepackage{algorithmic}

\setcopyright{popets}
\copyrightyear{YYYY}

\acmYear{YYYY}
\acmVolume{YYYY}
\acmNumber{X}
\acmDOI{XXXXXXX.XXXXXXX}
\acmISBN{}
\acmConference{Proceedings on Privacy Enhancing Technologies}
\newcommand{\ct}[1]{\mathbf{\underline{#1}}}
\newcommand{\cte}[1]{\mathbf{\underline{\tilde{#1}}}}
\newcommand{\cadd}{+_\bullet}
\newcommand{\padd}{+_\circ}
\newcommand{\csub}{-_\bullet}
\newcommand{\psub}{-_\circ}
\newcommand{\cmult}{\odot_\bullet}
\newcommand{\pmult}{\odot_\circ}
\newcommand{\crot}{\textit{Rot}_{\bullet}}
\newcommand{\csumc}{\sideset{_{}^{}}
   {_\bullet^{\text{cols}}}\sum}
\newcommand{\csumct}{\sum_\bullet^{\text{cols}}}
\newcommand{\csumr}{\sideset{_{}^{}}
   {_\bullet^{\text{rows}}}\sum}
\newcommand{\csumrt}{\sum_\bullet^{\text{rows}}}

\begin{document}

\title[ReBoot: Encrypted Training of Deep Neural Networks with CKKS Bootstrapping]{ReBoot: Encrypted Training of Deep Neural Networks \\ with CKKS Bootstrapping}


\author{Alberto Pirillo}
\orcid{}
\affiliation{%
  \institution{Politecnico di Milano}
  \city{Milan}
  \country{Italy}}
\email{alberto.pirillo@mail.polimi.it}

\author{Luca Colombo}
\orcid{0000-0002-1986-6459}
\affiliation{%
  \institution{Politecnico di Milano}
  \city{Milan}
  \country{Italy}}
\email{luca2.colombo@polimi.it}

\thanks{Preprint.}




\begin{abstract}
Growing concerns over data privacy underscore the need for deep learning methods capable of processing sensitive information without compromising confidentiality. Among privacy-enhancing technologies, Homomorphic Encryption (HE) stands out by providing post-quantum cryptographic security and end-to-end data protection, safeguarding data even during computation. While Deep Neural Networks (DNNs) have gained attention in HE settings, their use has largely been restricted to encrypted inference. Prior research on encrypted training has primarily focused on logistic regression or has relied on multi-party computation to enable model fine-tuning. This stems from the substantial computational overhead and algorithmic complexity involved in DNNs training under HE. In this paper, we present ReBoot, the first framework to enable fully encrypted and non-interactive training of DNNs. Built upon the CKKS scheme, ReBoot introduces a novel HE-compliant neural network architecture based on local error signals, specifically designed to minimize multiplicative depth and reduce noise accumulation. ReBoot employs a tailored packing strategy that leverages real-number arithmetic via \textit{SIMD} operations, significantly lowering both computational and memory overhead. Furthermore, by integrating approximate bootstrapping, ReBoot learning algorithm supports effective training of arbitrarily deep multi-layer perceptrons, making it well-suited for machine learning as-a-service. ReBoot is evaluated on both image recognition and tabular benchmarks, achieving accuracy comparable to 32-bit floating-point plaintext training while enabling fully encrypted training. It improves test accuracy by up to $+3.27\%$ over encrypted logistic regression, and up to $+6.83\%$ over existing encrypted DNN frameworks, while reducing training latency by up to $8.83\times$. ReBoot is made available to the scientific community as a public repository.
\end{abstract}


\maketitle


\section{Introduction}
\label{sec:introduction}


The increase concerns over data protection have sparked widespread interest in the integration of privacy enhancing technologies into Machine Learning (ML) and Deep Learning (DL) applications. Among these technologies, Homomorphic Encryption (HE) stands out for its unique ability to support computations directly on encrypted data~\citep{Acar2018Survey}. By offering post-quantum cryptographic security, HE enables data processing in untrusted environments, such as cloud services, without ever exposing sensitive information, thereby preserving confidentiality throughout computation~\citep{brakerski2013classical}. However, despite its potential, HE introduces significant challenges that must be addressed. For example, it typically supports only a limited subset of arithmetic operations on encrypted data, primarily additions and multiplications, which restricts computations to polynomial functions. As a result, common ML and DL operations, such as ReLU and sigmoid, are not supported in the encrypted domain. This limitation requires the use of polynomial approximations and the design of alternative model architectures specifically tailored for encrypted computation. Moreover, HE supports only a limited number of consecutive encrypted operations, as each operation introduces additional noise into the ciphertext. To preserve the integrity of encrypted data, a noise threshold is defined, beyond which the ciphertext can no longer be correctly decrypted.

To address this constraint, Fully Homomorphic Encryption (FHE) schemes employ bootstrapping~\citep{Gentry2009Fully}, a homomorphic operation that refreshes the ciphertext structure and reduces accumulated noise, thereby enabling unbounded computation. More recently, an \textit{approximate} variant of bootstrapping has been adapted for Leveled Homomorphic Encryption (LHE) schemes such as Cheon-Kim-Kim-Song (CKKS)~\citep{Cheon2017Homomorphic, Nir2022BLEACH}. Unlike in FHE schemes such as Brakerski--Fan--Vercauteren (BFV)~\citep{Fan2012Somewhat} and Torus FHE (TFHE)~\citep{Chilotti2018TFHE}, approximate bootstrapping does not enable unlimited computation, as it does not reduce accumulated noise~\citep{al2023demystifying}. Nevertheless, it significantly extends the number of encrypted operations that can be performed by restoring the ciphertext structure.

Bootstrapping plays a crucial role in enabling end-to-end encrypted learning. However, training ML models, and particularly DL models, remains a major open research challenge. This is mainly due to three factors: \textit{(i)} the inherent noise accumulation of HE, \textit{(ii)} the significant computational overhead of homomorphic operations, and \textit{(iii)} the functional limitations imposed by HE schemes. Consequently, existing research has largely focused on encrypted training of simpler ML models~\citep{Bergamaschi2019Homomorphic, Kim2018Logistic, Crockett2020Low, Han2018Efficient, Carpov2019Privacy}, or on fine-tuning the final layer of Deep Neural Networks (DNNs) within transfer learning frameworks~\citep{Walch2022Cryptotl, Zhang2022Privacy, Jin2020Secure, Al2020Privft, Lee2023Hetal, Panzade2024Icantifinetune}. More recent work has explored training entire DNNs over encrypted data by leveraging FHE schemes~\citep{Nandakumar2019Towards, Yoo2021tBMPNet, Montero2024Neural, Colombo2024Training, Lou2020Glyph}. However, some of these approaches rely on integer-based encryption, which necessitates complex learning algorithms that often compromise model accuracy~\citep{Pirillo2024Nitro}, while others are constrained by multi-party computation mechanisms~\citep{Mihara2020Neural}. To date, no prior work has addressed the challenge of training DNNs on encrypted data using the CKKS scheme in a fully non-interactive setting.


In this regard, the aim of this paper is to address the following research question: \emph{how can CKKS approximate bootstrapping be effectively leveraged to enable fully encrypted interaction-free training of DNNs?} To this end, we propose ReBoot, a novel encrypted learning framework specifically designed to train HE-compliant Multi-Layer Perceptrons (MLPs)~\citep{Bishop1995NeuralNF} on encrypted data. ReBoot components are specifically designed to manage noise accumulation while minimizing both computational and memory overhead associated with HE. To achieve this, ReBoot introduces a novel HE-compliant NN architecture along with an advanced packing technique that exploits CKKS ability to perform \emph{Single Instruction, Multiple Data (SIMD)} operations within ciphertexts. By integrating approximate bootstrapping, ReBoot learning algorithm supports deep encrypted computation pipelines, allowing fully non-interactive training of MLPs and making it well-suited for practical and scalable machine learning as-a-service scenarios. To the best of our knowledge, ReBoot is the first framework in the literature to extend CKKS-based training beyond logistic regression and fine-tuning, enabling end-to-end encrypted training of real-valued DNNs. ReBoot is implemented as an open-source Python library\footnote{The code will be released as a public repository in the next phase.} based on OpenFHE~\citep{Ahmad2022OpenFHE}.

In summary, our contributions are:
\begin{enumerate}
    \item The design of an advanced packing technique that minimizes both computational and memory overhead of HE by fully exploiting CKKS \textit{SIMD} capabilities.
    \item The design of a CKKS-compliant MLP architecture in which all operations are optimized to manage noise accumulation.
    \item The design of ReBoot encrypted learning algorithm, which enables non-interactive encrypted training of HE-compliant MLPs on encrypted data.
\end{enumerate}
Experimental results on state-of-the-art image recognition and tabular benchmarks demonstrate that ReBoot matches the accuracy of plaintext 32-bit floating-point (FP32) plaintext training, while enabling fully non-interactive end-to-end encrypted training. Moreover, it improves test accuracy by up to $+3.27\%$ over encrypted logistic regression, and up to $+6.83\%$ over existing encrypted DNN frameworks, while reducing training latency by up to $8.83\times$.

The paper is organized as follows. Section \ref{sec:related_literature} surveys related literature on encrypted training of ML and DL models. Section \ref{sec:background} introduces the fundamentals of the CKKS scheme. Section \ref{sec:proposed_solution} details our proposed ReBoot solution, while experimental results are presented in Section~\ref{sec:experimental_results}. Lastly, conclusions and future research directions are discussed in Section~\ref{sec:conclusions}.


\section{Related Literature}
\label{sec:related_literature}
This section provides an overview of HE literature on encrypted training of ML and DL models. Due to its architectural simplicity, logistic regression was among the first algorithms to be trained under encryption. Early studies presented proof-of-concept implementations using bootstrapping-free CKKS scheme~\citep{Bergamaschi2019Homomorphic, Kim2018Logistic, Crockett2020Low}. Subsequent research extended these works by integrating approximate bootstrapping into the training procedure, enabling more practical, real-world applications~\citep{Han2018Efficient, Carpov2019Privacy}. Building on these foundations, several studies have demonstrated the computational feasibility of encrypted fine-tuning as an effective method for applying transfer learning in privacy-critical contexts~\citep{Walch2022Cryptotl, Zhang2022Privacy, Jin2020Secure, Al2020Privft, Lee2023Hetal, Panzade2024Icantifinetune}. Rather than retraining entire networks, these approaches focus on updating only the final fully-connected layer of a pre-trained model, making them particularly well-suited for CKKS-based computation. 

More recent research has shifted toward training full DNNs over encrypted data by leveraging FHE schemes. To this end, several works have employed TFHE to develop learning algorithms that redesign small Multi-Layer Perceptrons (MLPs) to operate using encrypted Boolean logic gates~\citep{Yoo2021tBMPNet, Colombo2024Training, Montero2024Neural}. In particular, \citet{Montero2024Neural} explored interaction-based training, in which ciphertexts are \textit{re-encrypted} to refresh noise level after each mini-batch pass. \citet{Yoo2021tBMPNet} and \citet{Colombo2024Training} leveraged encrypted Boolean logic gates to implement commonly used MLP functions, as well as cross-validation procedures. By integrating bootstrapping, both works support non-interactive training of arbitrarily deep Neural Networks (NNs). However, the lack of \textit{SIMD} capabilities in the TFHE scheme forces homomorphic operations to be executed sequentially, resulting in prohibitive latency for practical DNN training. Consequently, these works remain at the proof-of-concept stage, demonstrating TFHE-based training only on toy-example architectures.

Other FHE schemes remain relatively underexplored.
For instance, \citet{Nandakumar2019Towards} investigated the use of the Brakerski–Gentry–Vaikuntanathan (BGV) scheme~\citep{Brakerski2012Leveled} to train small MLPs through a non-interactive learning algorithm that leverages BGV bootstrapping to refresh ciphertexts. Additionally, \citet{Lou2020Glyph} proposed a hybrid approach that combines the BFV and TFHE schemes to optimize training performance. Their method employs a BFV-based training algorithm for efficient arithmetic operations, while switching ciphertexts to the TFHE domain via \textit{key switching} to perform non-linear functions and bootstrapping. This design leverages the strengths of both schemes, thereby improving overall training times. Nonetheless, both BGV and BFV operate over integer values, necessitating the quantization of data and model parameters. This constraint requires the use of complex integer-based learning algorithms~\citep{Pirillo2024Nitro}, which not only increase design complexity but also degrade model accuracy compared to training in the real-valued domain.

To preserve full-precision weights during training while maintaining feasible latency, \citet{Mihara2020Neural} proposed a solution that leverages CKKS within an interaction-based protocol between the data owner (i.e., the client) and the service provider (i.e., the server). Their approach relies on a \textit{re-encryption} mechanism to both refresh noise in the model weights and re-order values within the ciphertexts. While effective in preserving model integrity, the solution is unsuitable for \textit{as-a-service} scenarios due to its reliance on frequent client-server interactions, which introduce significant communication overhead and operational costs.

In this context, ReBoot stands out as the first framework to enable fully encrypted non-interactive training of real-valued DNNs. Reboot introduces both a novel encrypted architecture and a corresponding encrypted learning algorithm that leverage the proposed packing technique to exploit CKKS \textit{SIMD} operations while addressing the scheme's inherent limitations, thereby reducing computational and memory demands.


\section{The CKKS Homomorphic Encryption Scheme }
\label{sec:background}

Homomorphic encryption (HE) is a class of encryption schemes that allows computations to be performed directly on encrypted data, while ensuring that the result obtained from computations on ciphertexts matches the result of performing the same operation on the corresponding plaintexts~\citep{Acar2018Survey}. Formally, an encryption function $E(\mathsf{pk}, \cdot)$ and its corresponding decryption function $D(\mathsf{sk}, \cdot)$ are considered homomorphic with respect to (w.r.t.) a set of functions $\mathcal{F}$ if, for any function $f \in \mathcal{F}$, there exists a function $g$ such that $f(x) = D\left(\mathsf{sk}, g\left( E(\mathsf{pk}, x) \right) \right)$ for any input $x$, where $\mathsf{pk}$ and $\mathsf{sk}$ denote the public and secret keys, respectively~\citep{Boemer2019nGraph}. This property relies on the preservation of the data's algebraic structure throughout the encryption and computation processes~\citep{ogburn2013homomorphic}.

Among the various HE schemes, our study employs CKKS~\citep{Cheon2017Homomorphic}. Specifically, CKKS is based on a widely used quantum-resistant computational problem known as the Ring Learning With Errors (RLWE) problem~\citep{Lyubashevsky2010OnIL, chase2017security}. It supports approximate arithmetic over encrypted complex numbers, making it well-suited for real-valued ML and DL applications. CKKS belongs to the family of \emph{leveled} HE schemes, which allow only a finite number of consecutive encrypted operations before the underlying plaintext becomes unrecoverable. This limit, known as \emph{scheme level} and denoted by $l$, arises from noise injected into ciphertexts to ensure the probabilistic security guarantees of the RLWE problem~\citep{goldwasser2019probabilistic}. The noise grows with each operation, eventually leading to an unrecoverable rounding error upon decryption.

The algebraic structure of CKKS plaintexts and ciphertexts is defined by a set of encryption parameters $\Theta = \{ N, \mathbf{q}, \Delta \}$, where $N$ is a power of two defining the \emph{polynomial degree}, $\mathbf{q}$ is a list of $l + 2$ prime numbers known as the \emph{coefficient moduli}, and $\Delta$ defines the \emph{scaling factor}. The parameters $\Theta$ determine the scheme's \emph{security level}, denoted by $\lambda$, as well as the encoding precision and the number of available operations. Each time a ciphertext is multiplied, its coefficient modulus is reduced by switching to the next prime in the modulus chain $\mathbf{q}$. This process continues until level $l$ is reached, beyond which further multiplications are not permitted due to excessive noise accumulation. Consequently, the choice of scheme level $l$ is critical, as it determines the maximum supported multiplicative depth and directly influences ciphertext size. A higher level $l$ requires a larger polynomial degree $N$, which, in turn, significantly increases both computational complexity and memory consumption~\citep{Albrecht2019Homomorphic}. Given the inherently complex nature of DNN training, minimizing the required multiplicative depth is essential to ensure efficient execution under encryption.

CKKS supports \emph{batching}~\citep{Smart2014Fully}, a technique that enables parallel processing through \emph{Single Instruction, Multiple Data (SIMD)} operations \citep{Flynn1966Very}. Using batching, a vector of up to $N/2$ values can be encoded into a single ciphertext, thereby reducing both memory and computational overhead. Formally, let $\mathbf{z} \in \mathbb{C}^{N/2}$ be a vector of complex numbers to be encrypted using CKKS. Before encryption, $\mathbf{z}$ is mapped to a polynomial $z(X) \in \mathcal{R}$ using a variant of the \emph{complex canonical embedding} map, denoted as $\phi : \mathbb{C}^{N/2} \rightarrow \mathcal{R}$, where $\mathcal{R} = \mathbb{Z}[X]/(X^N + 1)$ is the polynomial ring in which plaintexts reside. The polynomial $z(X)$ is then encrypted using the secret key $\mathsf{sk}$ to produce a ciphertext $\ct{z} \in \mathcal{R}_\mathbf{q} = \mathbb{Z}_{q_0}[X]/(X^N + 1)$. During encryption, a noise term is added to ensure the probabilistic nature of the encryption scheme, thereby preventing deterministic attacks.

The adopted CKKS scheme natively supports element-wise operations on encrypted vectors. Let $\ct{a} = [\underline{a}_1, \underline{a}_2, \dots, \underline{a}_{N/2}]$ and $\ct{b} = [\underline{b}_1, \underline{b}_2, \dots, \underline{b}_{N/2}]$ be two encrypted CKKS vectors. The element-wise homomorphic addition, subtraction, and multiplication, denoted by $\cadd, \csub$, and $\cmult$, respectively, are defined as follows:
\begin{align}
    & \ct{a} \cadd \ct{b} = \left[\underline{a}_1 + \underline{b}_1, \underline{a}_2 + \underline{b}_2, \dots, \underline{a}_{N/2} + \underline{b}_{N/2}\right], \label{eq:add} \\
    & \ct{a} \csub \ct{b} = \left[\underline{a}_1 - \underline{b}_1, \underline{a}_2 - \underline{b}_2, \dots, \underline{a}_{N/2} - \underline{b}_{N/2}\right], \label{eq:sub} \\
    & \ct{a} \cmult \ct{b} = \left[\underline{a}_1 \cdot \underline{b}_1, \underline{a}_2 \cdot \underline{b}_2, \dots, \underline{a}_{N/2} \cdot \underline{b}_{N/2}\right]. \label{eq:mult}
\end{align}
Similarly, element-wise addition, subtraction, and multiplication can be performed between a plaintext vector $\textbf{z}$ and an encrypted vector $\ct{a}$. In this case, the operations are denoted by $\padd, \psub$, and $\pmult$, respectively. The final operation supported by CKKS is homomorphic vector rotation, which enables a cyclic shift of the elements batched within a single ciphertext. A rotation by $k$ positions of the encrypted vector $\ct{a}$ results in a new encrypted vector whose elements are cyclically shifted as follows:
\begin{equation*}
    \crot\left(\ct{a}, k\right) = \left[\underline{a}_{k+1}, \underline{a}_{k+2}, \dots, \underline{a}_{k}\right].
\end{equation*}
Rotation keys are required to perform this operation, with one key corresponding to each desired rotation index $k$.

By leveraging batching, matrices can be encoded in their flattened form into a single ciphertext. This representation enables the definition of aggregate operations that compute homomorphic summations across specific dimensions of the original matrix~\citep{Han2018Efficient}. Consider a flattened encrypted CKKS matrix
\begin{equation*}
\ct{w} = \left[\underline{W}_{1,1}, \dots, \underline{W}_{1,c}, \underline{W}_{2,1}, \dots, \underline{W}_{2,c}, \dots, \underline{W}_{r,1}, \dots, \underline{W}_{r,c}\right]
\end{equation*}
of dimension $r \times c = N/2$, where both $r$ and $c$ are powers of two. The summation across rows is defined as
\begin{equation}\label{eq:sumrows}
\ct{s} = \csumr \ct{w},
\end{equation}
where
\begin{equation*}
\ct{s} = \left[\underline{s}_{1}, \dots, \underline{s}_{c}, \underline{s}_{1}, \dots, \underline{s}_{c}, \dots, \underline{s}_{1}, \dots, \underline{s}_{c}\right]
\end{equation*}
denotes a flattened encrypted matrix in which each value $\underline{s}_j$, with $j \in \{1, \dots, c\}$, is replicated $r$ times along the row dimension and represents the $j$-th column-wise sum of $\ct{w}$. More in detail, the summation across rows is computed by recursively performing homomorphic element-wise additions and rotations. Let $\ct{s}^0 = \ct{w}$ be the initial ciphertext. Then, for each iteration $t \in \{0, \ldots, \log_2(r) - 1\}$, the partial column-wise sums are computed as
\begin{equation*}
\ct{s}^{t+1} = \ct{s}^t \cadd \crot\left(\ct{s}^t, c \cdot 2^{\log_2(r) - 1 - t}\right).
\end{equation*}
After the final iteration, the ciphertext $\ct{s} = \ct{s}^{\log_2(r)}$ contains the replicated column-wise sums of $\ct{w}$.

Similarly, the summation across columns is defined as
\begin{equation}\label{eq:sumcols}
\ct{s} = \csumc \ct{w},
\end{equation}
where
\begin{equation*}
\ct{s} = \left[\underline{s}_{1}, \dots, \underline{s}_{1}, \underline{s}_{2}, \dots, \underline{s}_{2}, \dots, \underline{s}_{r}, \dots, \underline{s}_{r}\right]
\end{equation*}
denotes a flattened encrypted matrix in which each value $\underline{s}_i$, with $i \in \{1, \dots, r\}$, is replicated $c$ times along the column dimension and represents the $i$-th row-wise sum of $\ct{w}$. Specifically, the summation across columns involves three steps. First, the ciphertext accumulating the row-wise sums is computed recursively using homomorphic element-wise additions and rotations. Let $\ct{u}^0 = \ct{w}$ denote the initial ciphertext. Then, for each iteration $t \in \{0, \ldots, \log_2(c) - 1\}$, the partial row-wise sums are computed as
\begin{equation*}
\ct{u}^{t+1} = \ct{u}^t \cadd \crot\left(\ct{u}^t, 2^{\log_2(c) - 1 - t}\right).
\end{equation*}
After the final iteration, the ciphertext 
\begin{equation*}
\ct{u}^{\log_2(c)} = \left[\underline{u}_{1}, \bot ,\dots, \bot, \underline{u}_{2}, \bot, \dots, \bot, \dots, \underline{u}_{r}, \bot, \dots, \bot\right]
\end{equation*}
contains the $i$-th row-wise sum in the first slot of each row $i$, while the remaining $c - 1$ slots in each row contain corrupted values, denoted by $\bot$. These corrupted values result from additions with elements from subsequent rows introduced by rotation operations. Second, to retain only the values $\underline{u}_i$ in the first slot of each row within the ciphertext $\ct{u}^{\log_2(c)}$, with $i \in \{1, \dots, r\}$, a masking operation is applied:
\begin{equation*}
\ct{u} = \textbf{m} \pmult \ct{u}^{\log_2(c)},
\end{equation*}
where $\textbf{m} \in \{0,1\}^{N/2}$ is a flattened binary mask containing all zeros except for the first slot of each row, which is set to 1. It is worth noting that, unlike the summation across rows, the summation across columns requires a multiplicative depth of 1 due to this masking step. Third, to obtain the final ciphertext $\ct{s}$, the row-wise sums are replicated $c$ times along the column dimension by recursively performing homomorphic element-wise additions and rotations. Let $\ct{s}^0 = \ct{u}$ be the initial ciphertext. Then, for each iteration $t \in \{0, \ldots, \log_2(c) - 1\}$, the partially replicated sums are computed as
\begin{equation*}
\ct{s}^{t+1} = \ct{s}^t \cadd \crot\left(\ct{s}^t, -2^{t}\right).
\end{equation*}
After the final iteration, the ciphertext $\ct{s} = \ct{s}^{\log_2(c)}$ contains the replicated row-wise sums of $\ct{w}$.

To overcome the limitation on the number of consecutive multiplications and enable an arbitrary number of homomorphic operations, CKKS supports a procedure known as \emph{bootstrapping}~\citep{Gentry2009Fully}. Bootstrapping refreshes the ciphertext by homomorphically restoring its modulus level, thereby allowing additional homomorphic operations beyond the original scheme level $l$. Formally, the goal of the bootstrapping operation applied to a ciphertext $\ct{c}$ at level $l = 0$ is to compute a refreshed ciphertext $\ct{c}' = \text{\emph{Bootstrap}}\left(\ct{c}\right) $ at level $l' - d > 0$, where $d$ is the multiplicative depth of the bootstrapping circuit, such that $\ct{c}' \approx \ct{c}$. \citet{Cheon2018Bootstrapping} proposed the first procedure to compute this modular reduction in CKKS by homomorphically applying the encoding step, approximating the modular reduction via a scaled sine function, and then applying the decoding step\footnote{Refer to~\citet{Bossuat2021Efficient} and~\citet{Bae2022META-BTS} for an in-depth description of CKKS bootstrapping as used in this work.}. However, unlike other FHE schemes such as BGV~\citep{Brakerski2012Leveled} and TFHE~\citep{Chilotti2018TFHE}, CKKS bootstrapping does not refresh the noise accumulated in the ciphertext, due to the scheme’s approximate nature~\citep{Nir2022BLEACH}. As a result, even with bootstrapping, it remains impossible to reduce the noise in CKKS ciphertexts, requiring us to rethink and redesign both standard DNN architectures and training algorithms to natively take into account these constraints.


\section{The Proposed Solution} 
\label{sec:proposed_solution}
This section introduces the proposed ReBoot framework, specifically designed for HE-compliant encrypted training and inference of MLPs. Section~\ref{subsec:packing} presents the packing schemes designed to efficiently encode inputs, weights, activations, and gradients. The encrypted NN architecture is detailed in Section~\ref{subsec:architecture}, while Section~\ref{subsec:learning_algorithm} describes the encrypted learning algorithm. Finally, Section~\ref{subsec:depth} provides a comprehensive analysis of the multiplicative depth required by the ReBoot framework.

\subsection{Reboot Packing}
\label{subsec:packing}
Reboot encrypted processing is designed to maximize computational efficiency by employing a structured packing strategy within the ciphertext space. By carefully organizing data inside each ciphertext, our approach enables highly efficient encrypted computation, significantly enhancing the training of MLPs on encrypted data. 

ReBoot employs two distinct packing schemes to encode vectors and matrices into ciphertexts. In particular, inputs, weights, activations, and gradients are packed into matrices $\tilde{\textbf{M}} \in \mathbb{R}^{r \times c}$, with $r \times c = N/2$, and then encrypted into single ciphertexts $\cte{m}$. The encoding dimensions, namely $r$ and $c$, are determined by the DNN architecture, as explained in Section \ref{subsubsec:encrypted_blocks}, and remain consistent across all packed matrices. This packing structure is essential for defining \textit{a priori} a fixed set of CKKS rotation keys, which are required for ciphertext rotations, as explained in Section~\ref{sec:background}. Furthermore, it enables full exploitation of \textit{SIMD} operations on ciphertexts. By ensuring consistent and compact data representation, ReBoot packing reduces the number of ciphertext and minimizes the required homomorphic rotations and multiplications, thereby significantly lowering memory usage and computational overhead during encrypted model training.

Let $\mathbf{a} = [a_1, \dots, a_d] \in \mathbb{R}^{d}$ be a one-dimensional vector, with $d \leq N/2$. Reboot packing encodes the input vector $\mathbf{a}$ into a matrix $\tilde{\textbf{A}} \in \mathbb{R}^{r \times c}$. Specifically, the matrix $\tilde{\textbf{A}}$ can be structured in the \textit{repeated} format
\begin{equation*}
    \tilde{\textbf{A}}_{\text{rep}} = \begin{bmatrix}
    a_1 & {a}_2 & \dots & {a}_d & 0 & \dots & 0 \\
    {a}_1 & {a}_2 & \dots & {a}_d & 0 & \dots & 0 \\
    \vdots & \\
    {a}_1 & {a}_2 & \dots & {a}_d & 0 & \dots & 0 \\
    \end{bmatrix},
\end{equation*}
where each row $i \in \{1, \dots, r\}$ contains a copy of the vector $\mathbf{a}$, zero-padded to match the row length $c$. Alternatively, it can be structured in the \textit{expanded} format
\begin{equation*}
    \tilde{\textbf{A}}_{\text{exp}} = \begin{bmatrix}
    {a}_1 & {a}_1 & \dots & {a}_1 \\
    {a}_2 & {a}_2 & \dots & {a}_2 \\
    \vdots \\
    {a}_d & {a}_d & \dots & {a}_d \\
    0 & 0 & \dots & 0 \\
    \vdots \\
    0 & 0 & \dots & 0
    \end{bmatrix},
\end{equation*}
where each column $j \in \{1, \dots, c\}$ contains a copy of the vector $\mathbf{a}$, zero-padded to match the column length $r$. Both representations are padded with zeros to conform to the matrix dimensions $r \times c$.

Similarly, let 
\begin{equation*}
    \mathbf{W} = \begin{bmatrix}
    W_{11} & W_{12} & \cdots & W_{1k} \\
    W_{21} & W_{22} & \cdots & W_{2k} \\
    \vdots & & & 
    \\ W_{d1} & W_{d2} & \cdots & W_{dk} 
    \end{bmatrix} \in \mathbb{R}^{d \times k}
\end{equation*}
be a two-dimensional matrix, with $d \times k \leq N/2$. Reboot packing encodes the input matrix $\mathbf{W}$ into a matrix $\tilde{\textbf{W}} \in \mathbb{R}^{r \times c}$. Specifically, the matrix $\tilde{\textbf{W}}$ can be structured in the \textit{row-wise} encoded format
\begin{equation*}
    \tilde{\textbf{W}}_{\text{row}} = \begin{bmatrix}
    W_{11} & W_{12} & \cdots & W_{1k} & 0 & \dots & 0 \\
    W_{21} & W_{22} & \cdots & W_{2k} & 0 & \dots & 0 \\
    \vdots & & & 
    \\ W_{d1} & W_{d2} & \cdots & W_{dk} & 0 & \dots & 0 \\
    0 & 0 & \dots & 0 & 0 & \dots & 0 \\
    \vdots \\
    0 & 0 & \dots & 0 & 0 & \dots & 0 \\
    \end{bmatrix},
\end{equation*}
where the input matrix $\mathbf{W}$ is padded with $r - d$ rows and $c - k$ columns of zeros. Alternatively, the matrix $\tilde{\textbf{W}}$ can be structured in the \textit{column-wise} encoded format
\begin{equation*}
    \tilde{\textbf{W}}_{\text{col}} = \begin{bmatrix}
    W_{11} & W_{21} & \cdots & W_{d1} & 0 & \dots & 0 \\
    W_{12} & W_{22} & \cdots & W_{d2} & 0 & \dots & 0 \\
    \vdots & & & 
    \\ W_{1k} & W_{2k} & \cdots & W_{dk} & 0 & \dots & 0 \\
    0 & 0 & \dots & 0 & 0 & \dots & 0 \\
    \vdots \\
    0 & 0 & \dots & 0 & 0 & \dots & 0 \\
    \end{bmatrix},
\end{equation*}
where the transposed matrix $\mathbf{W}^\intercal$ is padded with $r - k$ rows and $c - d$ columns of zeros. Note that ReBoot packing employs zero-padding to maintain consistent encoding dimensions across ciphertexts.

The proposed packing strategy ensures full utilization of ciphertext slots, which is critical to maximize CKKS \textit{SIMD} capabilities during encrypted MLP training and inference. By exploiting these packing characteristics, ReBoot enables encrypted vector-matrix multiplications using only element-wise operations and eliminates the need for \textit{repacking}, a common procedure used to reorder values within ciphertexts. This approach minimizes both the number of homomorphic operations and the required multiplicative depth, thereby reducing the frequency of costly bootstrapping operations.

\subsection{ReBoot Encrypted Architecture}
\label{subsec:architecture}
This section presents the encrypted NN architecture of ReBoot, illustrated in Figure~\ref{fig:architecture}, along with its main components. Specifically, Section~\ref{subsubsec:enc_fc} introduces the encrypted fully-connected layer. The polynomial approximation used for the encrypted activation function is described in Section~\ref{subsubsec:encrypted_act}. Finally, Section~\ref{subsubsec:encrypted_blocks} details the encrypted local-loss block, where layers are stacked together.

The proposed ReBoot framework is designed for training encrypted MLPs on multi-class classification tasks defined over a dataset $\mathbf{X}=\left\{\mathbf{x}^{\mu},y^{\mu}\right\}_{\mu=1}^{Q}$, where $\mathbf{x}^{\mu}\in \mathbb{R}^{k_0}$ are the input samples of dimension $k_0$, $y^{\mu}\in \left\{1,\dots,o\right\}$ their corresponding class labels (with $o$ denoting the number of classes), and $Q$ is the size of the training set. Before training, each sample $\mathbf{x}^{\mu}$ and its corresponding one-hot class label $\mathbf{y}^{\mu}$ are first encoded using the proposed packing techniques into $\tilde{\textbf{X}}^\mu$ and $\tilde{\textbf{Y}}^\mu$, then flattened and encrypted into ciphertexts $\cte{x}^\mu$ and $\cte{y}^\mu$, respectively.

\subsubsection{Encrypted Fully-Connected Layer}
\label{subsubsec:enc_fc}
ReBoot encrypted fully-connected layer is designed to compute vector-matrix multiplication between an encrypted activation vector $\cte{a}$ and an encrypted weight matrix $\cte{w}$, resulting in an encrypted pre-activation vector $\cte{z}$. Specifically, it leverages ReBoot packing schemes, as detailed in the previous section, to efficiently perform \emph{SIMD} computations. The proposed encrypted vector-matrix multiplication can take two forms, depending on the dimension along which the summation is performed: \textit{(i)} Row Encrypted Matrix Multiplication (\textit{RE-Matmul}), and \textit{(ii)} Column Encrypted Matrix Multiplication (\textit{CE-Matmul}). For the ease of meaning, and without loss of generality, we will henceforth consider the non-flattened packed representations of the ciphertexts $\cte{a}$, $\cte{w}$, and $\cte{z}$, denoted as $\cte{A}$, $\cte{W}$, and $\cte{Z}$, respectively.

The \textit{RE-Matmul} operation implements the summation across rows $\csumrt$, as defined in Equation~\ref{eq:sumrows}, and is expressed as
\begin{equation}
\label{eq:enc_rematmul}
\cte{Z}_{\text{rep}} = \text{\emph{RE-Matmul}}\left(\cte{A}_{\text{exp}}, \cte{W}\right) = \csumr \cte{A}_{\text{exp}} \cmult \cte{W}, 
\end{equation}
where $\cmult$ denotes homomorphic element-wise multiplication defined in Equation~\ref{eq:mult}, $\cte{A}_{\text{exp}}$ and $\cte{Z}_{\text{rep}}$ represent the encrypted activation and pre-activation vectors in the \textit{expanded} and \textit{repeated} formats, respectively, and $\cte{W}$ is the encrypted weight matrix. As detailed in Section~\ref{sec:background}, summation across rows replicates each column-wise sum $r$ times along the row dimension, effectively resulting in the \textit{repeated} format. From a computational perspective, \textit{RE-Matmul} requires a multiplicative depth of 1, contributed by the element-wise multiplication, along with $\log_2(r)$ additions and rotations. Conversely, \textit{CE-Matmul} implements the summation across columns $\csumct$, as defined in Equation~\ref{eq:sumcols}, and is expressed as
\begin{equation}
\label{eq:enc_cematmul}
\cte{Z}_{\text{exp}} = \text{\emph{CE-Matmul}}\left(\cte{A}_{\text{rep}}, \cte{W}\right) = \csumc \cte{A}_{\text{rep}} \cmult \cte{W}, 
\end{equation}
where $\cte{A}_{\text{rep}}$ and $\cte{Z}_{\text{exp}}$ denote the encrypted activation and pre-activation vectors in the \textit{repeated} and \textit{expanded} formats, respectively. Summation across columns replicates each row-wise sum $c$ times along the column dimension, effectively resulting in the \textit{expanded} format. \textit{CE-Matmul} requires a multiplicative depth of 2, contributed by the element-wise multiplication and the masking step in the summation across columns operation, as explained in Section~\ref{sec:background}, along with $2 \, \log_2(c)$ additions and rotations.

Depending on the encoding of its weight matrix, ReBoot encrypted fully-connected layer can be categorized into two types: \textit{(i)} the Row Encrypted Fully-Connected layer (\textit{RE-FC}), whose weight matrix is in the \textit{row-wise} encoded format $\cte{W}_{\text{row}}$, and \textit{(ii)} the Column Encrypted Fully-Connected layer (\textit{CE-FC}), whose weight matrix is in the \textit{column-wise} encoded format $\cte{W}_{\text{col}}$. As explained in Section~\ref{subsubsec:forward}, during the forward pass, the \textit{RE-FC} layer performs a \textit{RE-Matmul} operation between the \textit{expanded} activation vector $\cte{A}_{\text{exp}}$ and the \textit{row-wise} encoded weight matrix $\cte{W}_{\text{row}}$. Conversely, the \textit{CE-FC} layer performs a \textit{CE-Matmul} operation between the \textit{repeated} activation vector $\cte{A}_{\text{rep}}$ and the \textit{column-wise} encoded weight matrix $\cte{W}_{\text{col}}$. As a result, the output of a \textit{RE-FC} layer is a pre-activation vector in the \textit{repeated} format, which aligns exactly with the input format expected by the \textit{CE-FC} layer, and vice versa. This alternation between output and input formats enables the composition of consecutive encrypted layers without requiring explicit repacking of intermediate ciphertexts, as detailed in Section~\ref{subsubsec:encrypted_blocks}. Thanks to \textit{RE-FC} and \textit{CE-FC} layers, ReBoot can implement efficient and scalable inference and training of MLPs.

Let $\mathcal{H} = \{1, \dots, H\}$ denote the set of ReBoot encrypted fully-connected layers in the considered MLP, where $H$ is the total number of hidden layers. Let $k_h \in \mathbb{N}$ denote the dimensionality of layer $h \in \mathcal{H}$, with $k_0$ representing the input dimension of the encrypted MLP. The set $\mathcal{H}$ can be partitioned into two disjoint subsets: \textit{(i)} the set of \textit{RE-FC} layers, defined as $\mathcal{R} = \left\{ h \in \mathcal{H} \mid h \bmod 2 = 1 \right\}$, corresponding to odd-indexed layers, and \textit{(ii)} the set of \textit{CE-FC} layers, defined as $\mathcal{C} = \{ h \in \mathcal{H} \mid h \bmod 2 = 0 \}$, corresponding to even-indexed layers. The encoding dimensions of inputs, weights, activations, and gradients used in ReBoot packing, namely $r$ and $c$, as explained in Section~\ref{subsec:packing}, are defined as follows:
\begin{flalign*}
\begin{cases}
    r =  2^{\left\lceil \log_2 \left(\max \left( k_0, \max_{h \in \mathcal{R}} k_h \right)\right)\right\rceil} \\
    c =  2^{\left\lceil \log_2(  \max_{h \in \mathcal{C}} k_h) \right\rceil}
\end{cases},
\end{flalign*}
where $\lceil\cdot \rceil$ denotes the ceiling operation.

\subsubsection{Encrypted Activation Function}
\label{subsubsec:encrypted_act}
ReBoot encrypted activation function is designed to apply an encrypted non-linearity to the encrypted pre-activation vector $\cte{Z}$, producing the encrypted activation vector $\cte{A}$ as output. However, due to the limited set of operations supported by CKKS, commonly used activation functions in MLP models, such as ReLU, tanh, and sigmoid, cannot be directly applied in the encrypted domain. To address this limitation, ReBoot leverages \emph{EncryptedPolyReLU}, a second-degree polynomial approximation of the \emph{ReLU} function, specifically designed to be HE-compliant~\citep{Agarap2019Deep, Ali2024Polynomial}. \emph{EncryptedPolyReLU} is defined as follows:
\begin{equation}\label{eq:enc_polyrelu}
\cte{A} = \left(\cte{Z} \cmult \cte{Z}\right) \cadd \cte{Z},
\end{equation}
where $\cadd$ and $\cmult$ denote homomorphic element-wise addition and multiplication, respectively, as defined in Equation~\ref{eq:add} and Equation~\ref{eq:mult}. From a computational perspective, \textit{EncryptedPolyReLU} has a multiplicative depth of 1, contributed by the element-wise multiplication, along with a single addition. Since only element-wise operations are involved, \textit{EncryptedPolyReLU} does not alter the encoding structure between input and output encrypted vectors. Accordingly, if the input is in the \textit{expanded} format, the output will also be in the \textit{expanded} format, and vice versa.

\begin{figure}[t]
    \centering
    \includegraphics[width=\linewidth]{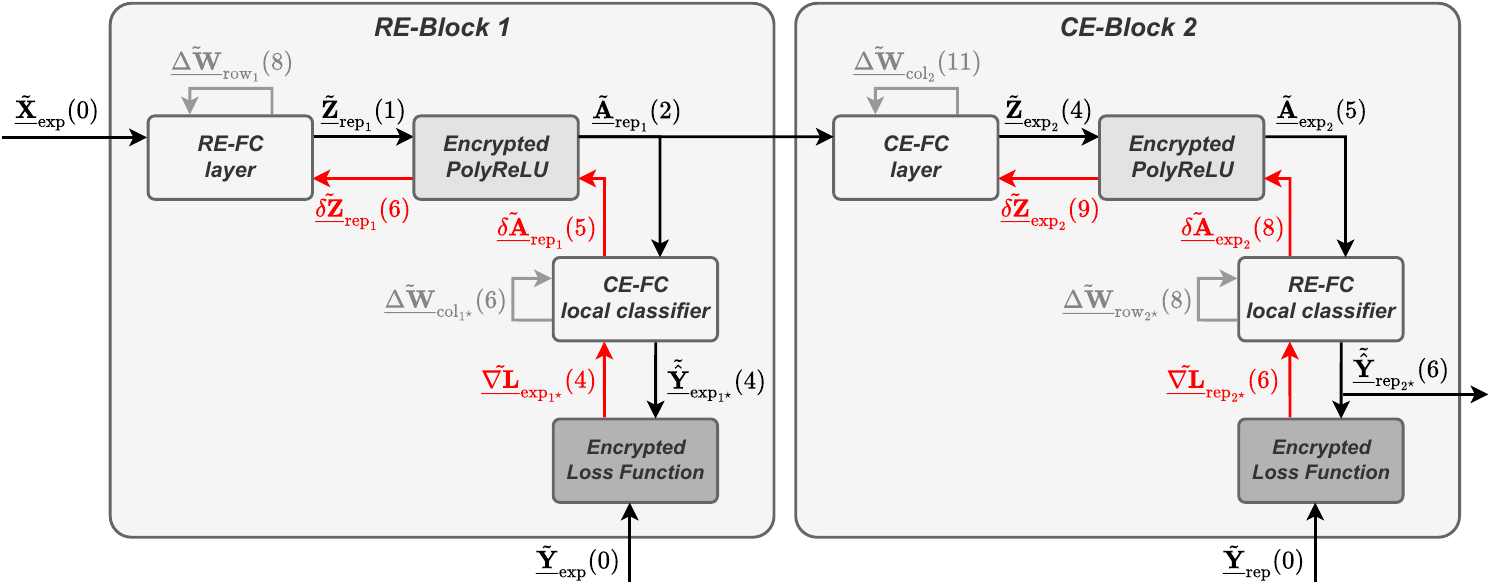}
    \caption{Overview of ReBoot encrypted architecture for a MLP with two hidden layers. The contributions of the \emph{forward pass}, \emph{backward pass}, and \emph{weight updates} are shown in black, red, and gray, respectively. The multiplicative depth of each value is indicated in parentheses.}
    \label{fig:architecture}
\end{figure}

\subsubsection{Encrypted Local-Loss Block}
\label{subsubsec:encrypted_blocks}
To minimize multiplicative depth required for encrypted training, and consequently the frequency of costly bootstrapping operations, Reboot encrypted architecture segments the MLP into multiple encrypted local-loss blocks. This design aligns with a stream of literature that employs local layer-wise loss functions~\citep{mostafa2018deep, Nøkland2019Training, patel2023local}. The key advantage of ReBoot architecture lies in its ability to confine gradient propagation within each encrypted local-loss block, thereby eliminating the need to back-propagate gradients through the entire NN. As a result, encrypted local-loss blocks restrict HE noise accumulation during training. Moreover, they enable the independent training of each block, since no information flows between blocks during the backward pass. This property allows for the parallelization of the training procedure, significantly reducing computational overhead. A detailed analysis of ReBoot's multiplicative depth is provided in Section~\ref{subsec:depth}.

ReBoot encrypted architecture is illustrated in Figure \ref{fig:architecture}. Each ReBoot encrypted fully-connected layer $h \in \mathcal{H}$, along with its associated \textit{EncryptedPolyReLU}, is paired with a local classifier $h^\star$ of size $k_h \times o$, forming an encrypted local-loss block $B_h$. The local classifier $h^\star$, which is itself a ReBoot encrypted fully-connected layer, serves to reduce the dimensionality of the encrypted activation vector $\cte{A}_h$ to the output dimension $o$. During the backward pass, the local output $\cte{\hat{Y}}_{h}$ and the true label $\cte{Y}$ are used by the local loss function $\text{L}_h$ to compute the local gradient $\cte{\nabla L}_h$. This gradient is then used by ReBoot training algorithm, detailed in Section~\ref{subsec:learning_algorithm}, to update the weights of both the encrypted fully-connected layer $h$ and its local classifier $h^\star$. In the final block $B_H$, the local classifier $H^\star$ serves as the NN's output layer, producing the final prediction $\cte{\mathbf{Y'}}_H$. Once the training process is completed, the intermediate local classifiers can either be discarded or leveraged for early-exit strategies~\cite{teerapittayanon2016branchynet, scardapane2020should, casale2023scheduling}.

Encrypted local blocks $B_h$ can be of two types: \textit{Row Encrypted Blocks} (\emph{RE-Block}s), when $h \in \mathcal{R}$, and \textit{Column Encrypted Blocks} (\emph{CE-Block}s), when $h \in \mathcal{C}$. It is worth noting that a \textit{RE-Block} has a local classifier $h^\star \in \mathcal{C}$, whereas a \textit{CE-Block} has a local classifier $h^\star \in \mathcal{R}$. By structuring ReBoot encrypted architecture to alternate between \textit{RE-Block}s and \textit{CE-Block}s, compatibility between encrypted fully-connected layers is inherently preserved, eliminating the need for auxiliary transformations. This property offers a significant computational advantage, as homomorphic repacking operations are typically expensive in terms of multiplicative depth, a critical factor for minimizing noise growth and avoiding costly bootstrapping operations. Consequently, alternating between \textit{RE-Block}s and \textit{CE-Block}s not only preserves packing efficiency across layers but also enables modular and scalable encrypted model architectures, making ReBoot particularly well-suited for practical encrypted MLP training and inference.

\subsection{ReBoot Encrypted Learning Algorithm}
\label{subsec:learning_algorithm}
This section outlines the key components of ReBoot encrypted learning algorithm, as detailed in Algorithm~\ref{alg:learning_alg} and Algorithm~\ref{alg:weight_updates}. Since each encrypted local-loss block is trained independently, as explained in Section~\ref{subsubsec:encrypted_blocks}, and the extension to \emph{CE-Block}s is straightforward, we focus on describing the training procedure for a representative \emph{RE-Block} $B_h$, with $h \in \mathcal{R}$. Specifically, Section~\ref{subsubsec:forward} describes the forward pass of a \emph{RE-Block}. Section~\ref{subsubsec:encrypted_loss} introduces the encrypted loss function along with its gradient computation. In Section~\ref{subsubsec:backward}, the backward pass within the \emph{RE-Block} is presented. Finally, Section~\ref{subsubsec:encrypted_updates} details the encrypted weight update process.

\subsubsection{Forward Pass}
\label{subsubsec:forward}
The forward pass procedure is executed at the beginning of each iteration $t \in \{1, \dots, T\}$, where $T$ denotes the total number of training iterations. Let $\cte{A}_{\text{exp}_{h-1}}^t$ be the encrypted activations produced by the previous \emph{CE-Block} $B_{h-1}$. Notably, in the first local-loss block (i.e., when $h = 1$), the input activations correspond to the input training samples, i.e., $\cte{A}_{\text{exp}_{0}}^t = \cte{X}_{\text{exp}}^t$.
The \emph{RE-Block} computes the encrypted pre-activations as
\begin{equation*}
    \cte{Z}_{\text{rep}_h}^t = \text{\emph{RE-Matmul}}\left(\cte{A}_{\text{exp}_{h-1}}^t, \cte{W}_{\text{row}_{h}}^t\right),
\end{equation*}
where $\cte{W}_{\text{row}_{h}}^t$ denotes the encrypted weights of the \emph{RE-FC} layer $h$ in the \textit{row-wise} encoded format, and \emph{RE-MatMul} refers to the row encrypted matrix multiplication described in Equation~\ref{eq:enc_rematmul}. The resulting encrypted pre-activations $\cte{Z}_{\text{rep}_h}^t$ are then passed through the \emph{EncryptedPolyReLU} activation function described in Equation~\ref{eq:enc_polyrelu}, yielding the encrypted activations $\cte{A}_{\text{rep}_h}^t$. These activations are simultaneously propagated to both the next \emph{CE-Block} $B_{h+1}$ and the corresponding local classifier $h^\star$. Specifically, the local classifier produces the encrypted predictions
\begin{equation*}
    \cte{\hat{Y}}_{\text{exp}_h\star}^t = \text{\emph{CE-Matmul}}\left(\cte{A}_{\text{rep}_{h}}^t, \cte{W}_{\text{col}_{h^\star}}^t\right),
\end{equation*}
where $\cte{W}_{\text{col}_{h^\star}}^t$ denotes the encrypted weights of the \emph{CE-FC} local classifier $h^\star$ in the \textit{column-wise} encoded format, and \emph{CE-Matmul} refers to the column encrypted matrix multiplication defined in Equation~\ref{eq:enc_cematmul}. The local predictions $\cte{\hat{Y}}_{\text{exp}_h\star}^t$ are then used by the encrypted loss function, described in the following section, to compute the gradients required to update the weights of both the \emph{RE-FC} layer $h$ and the \emph{CE-FC} local classifier $h^\star$. This design mitigates ciphertext noise accumulation by limiting multiplicative depth, as analyzed in Section~\ref{subsec:depth}.

\begin{algorithm}[!t]
    \small
    \caption{ReBoot encrypted learning algorithm: \emph{RE-Block} $B_h$}
    \label{alg:learning_alg}
    \begin{algorithmic}[1]
        \REQUIRE{$\cte{W}^t_{\text{row}_{h}}$: encrypted layer weights at iteration $t$, $\cte{W}^t_{\text{col}_{h^\star}}$: encrypted local classifier weights at iteration $t$, $\cte{A}^t_{\text{exp}_{h-1}}$: encrypted activations of block $B_{h-1}$ at iteration $t$, $\gamma$: learning rate, $\eta$: decay rate, $\mu$: momentum, $t$: iteration}
        \ENSURE{$\cte{W}^{t+1}_{\text{row}_{h}}$: encrypted layer weights at iteration $t+1$, $\cte{W}^{t+1}_{\text{col}_{h^\star}}$: encrypted local classifier weights at iteration $t+1$}
    
        \vspace{0.2cm}
        // Forward Pass
        \STATE{$\cte{Z}^t_{\text{rep}_h} \leftarrow  \text{\emph{RE-Matmul}}\left(\cte{A}^t_{\text{exp}_{h-1}}, \cte{W}^t_{\text{row}_{h}}\right)$}
        
        \STATE{$\cte{A}^t_{\text{rep}_h} \leftarrow  \text{\emph{EncryptedPolyReLU}}\left(\cte{Z}^t_{\text{rep}_h}\right)$}

        \STATE{$\cte{\hat{Y}}^t_{\text{exp}_h} \leftarrow  \text{\emph{CE-Matmul}}\left(\cte{A}^t_{\text{rep}_{h}}, \cte{W}^t_{\text{col}_{h^\star}}\right)$}

        \vspace{0.2cm}
        // Encrypted Gradient of Loss Function
        \STATE{$\cte{\nabla L}^t_{\text{exp}_h} \leftarrow  \cte{\hat{Y}}^t_{\text{exp}_h} \csub \cte{Y}^t_{\text{exp}_h}$}

        \vspace{0.2cm}
        // Backward Pass
        \STATE{$\cte{\delta W}^t_{\text{col}_{h^\star}} \leftarrow  \cte{A}^t_{\text{rep}_{h}} \cmult \cte{\nabla L}^t_{\text{exp}_h}$}

        \STATE{$\cte{\delta A}^t_{\text{rep}_h} \leftarrow  \text{\emph{RE-Matmul}}\left(\cte{\nabla L}^t_{\text{exp}_h}, \cte{W}^t_{\text{col}_{h^\star}}\right)$}

        \STATE{$\cte{\delta Z}^t_{\text{rep}_h} \leftarrow \text{\emph{EncryptedPolyReLU}}^{\, \prime} \left(\cte{\delta A}^t_{\text{rep}_h}\right)$}

        \STATE{$\cte{\delta W}^t_{\text{row}_{h}} \leftarrow  \cte{A}^t_{\text{exp}_{h-1}} \cmult \cte{\delta Z}^t_{\text{rep}_h}$}


        \vspace{0.2cm}
        // Weight Updates
        \STATE{$\cte{W}^{t+1}_{\text{row}_{h}}, \cte{V}^{t+1}_{\text{row}_{h}} \leftarrow  \text{\emph{UpdateWeights}}\left(\cte{W}^t_{\text{row}_{h}}, \cte{V}^t_{\text{row}_{h}}, \cte{\delta W}^t_{\text{row}_{h}}, \gamma, \eta, \mu, t \right)$}

        \STATE{$\cte{W}^{t+1}_{\text{col}_{h^\star}}, \cte{V}^{t+1}_{\text{col}_{h^\star}} \leftarrow  \text{\emph{UpdateWeights}}\left(\cte{W}^t_{\text{col}_{h^\star}}, \cte{V}^t_{\text{col}_{h^\star}}, \cte{\delta W}^{t-1}_{\text{col}_{h^\star}}, \gamma, \eta, \mu, t \right)$}
        
        \STATE{$\left\{\cte{W}^{t+1}_{\text{row}_{h}}, \cte{V}^{t+1}_{\text{row}_{h}} \right\} \leftarrow  \text{\emph{Bootstrap}}\left(\left\{\cte{W}^{t+1}_{\text{row}_{h}}, \cte{V}^{t+1}_{\text{row}_{h}} \right\}\right)$}

        \STATE{$\left\{\cte{W}^{t+1}_{\text{col}_{h^\star}}, \cte{V}^{t+1}_{\text{col}_{h^\star}} \right\} \leftarrow  \text{\emph{Bootstrap}}\left(\left\{\cte{W}^{t+1}_{\text{col}_{h^\star}}, \cte{V}^{t+1}_{\text{col}_{h^\star}} \right\}\right)$}

        \vspace{0.15cm}
        \RETURN{$\cte{W}^{t+1}_{\text{row}_{h}}, \cte{W}^{t+1}_{\text{col}_{h^\star}}$}
    \end{algorithmic}
\end{algorithm}

\subsubsection{Encrypted Loss Function}
\label{subsubsec:encrypted_loss}
The objective of ReBoot encrypted learning algorithm is to minimize the encrypted local loss function $\text{L}_h$ at each local block $B_h$. Specifically, ReBoot employs the \textit{Residual Sum of Squares (RSS)} loss, defined as
\begin{equation*}
    \cte{L}_{\text{exp}_h\star}^t = (\cte{\hat{Y}}_{\text{exp}_h\star}^t \csub \cte{Y}_{\text{exp}_h\star}^t) \cmult (\cte{\hat{Y}}_{\text{exp}_h\star}^t \csub \cte{Y}_{\text{exp}_h\star}^t),
\end{equation*}
where $\cte{\hat{Y}}_{\text{exp}_h\star}^t$ denotes the encrypted predictions and $\cte{Y}_{\text{exp}_h\star}^t$ the encrypted one-hot target labels. Here, $\csub$ and $\cmult$ represent homomorphic element-wise subtraction and multiplication, as defined in Equation~\ref{eq:sub} and Equation~\ref{eq:mult}, respectively.

The encrypted gradient of the loss w.r.t. the predictions, used to update the weights during the backward pass, is given by
\begin{equation*}
    \cte{\nabla L}_{\text{exp}_h\star}^t = \cte{\hat{Y}}_{\text{exp}_h\star}^t \csub \cte{Y}_{\text{exp}_h\star}^t.
\end{equation*}
The \textit{RSS} gradient offers two key advantages in the encrypted domain. First, it does not increase the multiplicative depth, thereby limiting noise accumulation and improving runtime efficiency. Second, it coincides with the gradient of the \textit{cross-entropy} loss when coupled with a \textit{sigmoid} activation, allowing the model to retain meaningful learning dynamics while remaining compatible with CKKS constraints.

\subsubsection{Backward Pass}
\label{subsubsec:backward}
Once the encrypted gradient of the loss w.r.t. the predictions has been computed, the corresponding encrypted gradients w.r.t. weights, activations, and pre-activations can be derived. First, the encrypted gradient w.r.t. the weights of the \emph{CE-FC} local classifier $\cte{\delta W}^t_{\text{col}_{h\star}}$ is computed as the outer product between the input activations to the local classifier $\cte{A}^t_{\text{rep}_{h}}$ and the loss gradient w.r.t. the predictions $\cte{\nabla L}^t_{\text{exp}_h\star}$. In the encrypted domain, thanks to ReBoot packing, this outer product is mapped to an element-wise multiplication $\cmult$ between two vectors encoded in the \textit{repeated} and \textit{expanded} formats, respectively. Formally,
\begin{equation*}
    \cte{\delta W}^t_{\text{col}_{h\star}} = \cte{A}^t_{\text{rep}_{h}} \cmult \cte{\nabla L}^t_{\text{exp}_h\star}.
\end{equation*}

Second, the encrypted gradient w.r.t. the activations $\cte{\delta A}^t_{\text{rep}_h}$ is computed as the matrix multiplication between the loss gradient w.r.t. the predictions $\cte{\nabla L}^t_{\text{exp}_h\star}$ and the transposed weight matrix $\cte{W}^t_{\text{col}_{h^\star}}$. Thanks to ReBoot packing, this operation can be efficiently performed using the row encrypted matrix multiplication \emph{RE-Matmul}, as defined in Equation~\ref{eq:enc_rematmul}, provided that the weights are encoded in the \textit{column-wise} format. Indeed, as explained in Section~\ref{subsec:packing}, the \textit{column-wise} encoding inherently represents the transposed weight matrix. Consequently, no additional operations are required to reorder values within the ciphertext. Formally,
\begin{equation*}
    \cte{\delta A}^t_{\text{rep}_h} = \text{\emph{RE-Matmul}}\left(\cte{\nabla L}^t_{\text{exp}_h}, \cte{W}^t_{\text{col}_{h^\star}}\right).
\end{equation*}

Third, the encrypted gradient w.r.t. the pre-activations $\cte{\delta Z}^t_{\text{rep}_h}$ is obtained by evaluating the homomorphic derivative of the \emph{EncryptedPolyReLU} activation function, defined in Equation~\ref{eq:enc_polyrelu}, at the encrypted activations $\cte{\delta A}^t_{\text{rep}_h}$. Specifically, the derivative \emph{EncryptedPolyReLU}$^{\, \prime}$ is computed as
\begin{equation*}
    \cte{\delta Z}^t_{\text{rep}_h} = \textbf{1}_{r \times c} \padd \textbf{2}_{r \times c} \pmult \cte{\delta A}^t_{\text{rep}_h},
\end{equation*}
where $\left\{\textbf{1}_{r \times c}, \textbf{2}_{r \times c}\right\} \in \mathbb{R}^{r \times c}$ are plaintext matrices filled with the constants 1 and 2, respectively, to match the dimensions of the encrypted activations. Here, $\padd$ and $\pmult$ represent plaintext-ciphertext element-wise addition and multiplication, respectively.

Lastly, the encrypted gradient w.r.t. the weights of the \emph{RE-FC} layer $\cte{\delta W}^t_{\text{row}_{h}}$ is obtained as
\begin{equation*}
    \cte{\delta W}^t_{\text{row}_{h}} = \cte{A}^t_{\text{exp}_{h-1}} \cmult \cte{\delta Z}^t_{\text{rep}_h}.
\end{equation*}
At this point, ReBoot encrypted weight update algorithm is executed to update the weights of both the \emph{CE-FC} local classifier $h^\star$ and the \emph{RE-FC} layer $h$, as explained in the following section.

\subsubsection{Encrypted Weight Update}
\label{subsubsec:encrypted_updates}
Once the encrypted gradients of the loss w.r.t. the weights of the \emph{CE-FC} local classifier $h^\star$ and the \emph{RE-FC} layer $h$ have been computed, they are used by ReBoot encrypted weight update algorithm, outlined in Algorithm~\ref{alg:weight_updates}, to update the corresponding weights. Given the incompatibility of most commonly used adaptive learning rate algorithms with HE, our method draws inspiration from \emph{Nesterov Accelerated Gradient (NAG)}~\citep{Sutskever2013Importance} to support both \textit{weight decay} and \textit{momentum} within the encrypted domain. This choice is motivated by the importance of these techniques in modulating convergence speed, which is especially important in HE-based training, where each homomorphic operation introduces additional noise and computational overhead.

Since the update procedure is general and applies to both \emph{CE-FC} and \emph{RE-FC} layers, we generically denote the encrypted gradient w.r.t. the weights as $\cte{\delta W}^t$. Let $\eta$ denote the \textit{weight decay rate}, $\mu$ the \textit{momentum}, and $\gamma$ the \textit{learning rate}. The encrypted decay-regularized gradient is computed as
\begin{equation*}
    \cte{\delta W}^{t} = \cte{\delta W}^t \cadd \left( \boldsymbol{\eta}_{r \times c} \pmult \cte{\delta W}^t \right),
\end{equation*}
where $\boldsymbol{\eta}_{r \times c} \in \mathbb{R}^{r \times c}$ is a plaintext matrix filled with the constant $\eta$ to match the dimensions of the encrypted weight gradient. Here, $\cadd$ and $\pmult$ represent homomorphic element-wise addition and plaintext-ciphertext element-wise multiplication, respectively.

The encrypted \textit{velocity} accumulates gradient information over time, serving as a momentum term that smooths updates and accelerates convergence. At the first iteration (i.e., $t = 1$), the encrypted velocity is initialized as
\begin{equation*}
    \cte{V}^{2} = \cte{\delta W}^{1},
\end{equation*}
and the corresponding weight update is given by
\begin{equation*}
    \cte{\Delta W}^{1} = \left(\boldsymbol{\gamma}_{r \times c} \pmult \cte{\delta W}^{1} \right) \cadd \left( \boldsymbol{\gamma\mu}_{r \times c} \pmult \cte{\delta W}^{1} \right).
\end{equation*}
For subsequent iterations $t \in \{2, \dots, T\}$, the encrypted velocity is updated as
\begin{equation*}
    \cte{V}^{t+1} = \left(\boldsymbol{\mu}_{r \times c} \pmult \cte{V}^{t} \right)  \cadd \cte{\delta W}^{t}.
\end{equation*}
Consequently, the encrypted weight update becomes
\begin{equation*}
    \cte{\Delta W}^{t} = \left(\boldsymbol{\gamma}_{r \times c} \pmult \cte{\delta W}^{t} \right) \cadd \left(\boldsymbol{\gamma\mu}_{r \times c} \pmult \cte{\delta W}^{t} \right) \cadd \left(\boldsymbol{\gamma\mu^2}_{r \times c} \pmult \cte{V}^{t} \right),
\end{equation*}
where $\left\{\boldsymbol{\gamma}_{r \times c}, \boldsymbol{\gamma\mu}_{r \times c}, \boldsymbol{\gamma\mu^2}_{r \times c}\right\} \in \mathbb{R}^{r \times c}$ are plaintext matrices filled with the constants $\gamma$, $\gamma\mu$, and $\gamma\mu^2$, respectively, to match the dimensions of the encrypted weight gradient and velocity. Finally, the encrypted weights are updated as
\begin{equation*}
    \cte{W}^{t+1} = \cte{W}^{t} \csub \cte{\Delta W}^{t},
\end{equation*}
where $\csub$ denotes homomorphic element-wise subtraction, as defined in Equation~\ref{eq:sub}.

At the end of ReBoot encrypted weight update procedure, two iterations of \textit{approximate bootstrapping} are performed to refresh the ciphertext modulus \textbf{q} of weights and velocities associated with both the \emph{RE-FC} layer $h$ and the \emph{CE-FC} local classifier $h^\star$. This step enables further encrypted operations beyond the original scheme level $l$, as explained in Section~\ref{sec:background}.

\begin{algorithm}[t]
\small
    \caption{\emph{UpdateWeights}: ReBoot encrypted weight update}
    \label{alg:weight_updates}
    \begin{algorithmic}[1]
        \REQUIRE{$\cte{W}^t$: encrypted layer weights at iteration $t$, $\cte{V}^t$: encrypted velocity at iteration $t$, $\cte{\delta W}^t$: encrypted weights gradient at iteration $t$, $\gamma$: learning rate, $\eta$: decay rate, $\mu$: momentum, $t$: iteration}
        \ENSURE{$\cte{W}^{t+1}$: encrypted layer weights at iteration $t+1$, $\cte{V}^{t+1}$: encrypted velocity at iteration $t+1$}
        
        \vspace{0.2cm}
        // Decay Regularization
        \STATE{$\cte{\delta W}^{t} = \cte{\delta W}^t \cadd \left( \boldsymbol{\eta}_{r \times c} \pmult \cte{\delta W}^t \right)$}

        \vspace{0.2cm}
        // NAG Algorithm
        \IF{$t=1$}

            \STATE{$\cte{V}^{2} \leftarrow  \cte{\delta W}^{1}$}
        
            \STATE{$\cte{\Delta W}^{1} = \left(\boldsymbol{\gamma}_{r \times c} \pmult \cte{\delta W}^{1} \right) \cadd \left( \boldsymbol{\gamma\mu}_{r \times c} \pmult \cte{\delta W}^{1} \right)$}
            
        \ELSE

            \STATE{$\cte{V}^{t+1} = \left(\boldsymbol{\mu}_{r \times c} \pmult \cte{V}^{t} \right)  \cadd \cte{\delta W}^{t}$}
            
            \STATE{$\cte{\Delta W}^{t} = \left(\boldsymbol{\gamma}_{r \times c} \pmult \cte{\delta W}^{t} \right) \cadd \left(\boldsymbol{\gamma\mu}_{r \times c} \pmult \cte{\delta W}^{t} \right) \cadd \left(\boldsymbol{\gamma\mu^2}_{r \times c} \pmult \cte{V}^{t} \right)$}
            
        \ENDIF
        \STATE{\vspace{0.1cm}$\cte{W}^{t+1} \leftarrow  \cte{W}^{t} \csub \cte{\Delta W}^t$}

        \vspace{0.15cm}
        \RETURN{\vspace{0.1cm}$\cte{W}^{t+1},  \cte{V}^{t+1}$}
    \end{algorithmic}
\end{algorithm} 

\begin{table}[t]
  \footnotesize
  \caption{Maximum multiplicative depth $\tau$ for standard BP and ReBoot. $H$ is the number of hidden layers in the MLP.}
  \label{tab:depth}
  \centering
  \begin{tabular}{ccc}
    \toprule
    Algorithm & Forward pass $\tau^{\text{fw}}$ & Backward pass $\tau^{\text{bw}}$\\
    \midrule
    BP      & $\lfloor 2.5\cdot H \rfloor + \lfloor 1.5 + (H \mod 2) \rfloor$ & $\lfloor 2.5\cdot H \rfloor +2$\\
    ReBoot  & $\lfloor 2.5 \cdot H \rfloor + \lfloor 1.5 + (H \mod 2) \rfloor$ & $\lfloor 4.5 + 1 -  (H \mod 2) \rfloor$ \\
 \bottomrule
  \end{tabular}
\end{table}

\subsection{ReBoot Multiplicative Depth}
\label{subsec:depth}
To correctly set CKKS encryption parameters, as explained in Section~\ref{sec:background}, it is essential to determine the maximum number of consecutive homomorphic multiplications required per training iteration, referred to as the multiplicative depth. Specifically, the multiplicative depth, denoted by $\tau$, depends on the network architecture and the learning algorithm, and can be expressed as a function of the number of hidden layers $H$ in the considered MLP. Minimizing the multiplicative depth is critical, as it directly impacts computational and memory efficiency. In this context, to highlight ReBoot's efficiency, we compare the multiplicative depth required to train an encrypted MLP using our proposed framework with that required by the standard Back-Propagation (BP) algorithm. 

In standard BP, encrypted gradients are propagated backward from the output classifier to the input layer. As a result, the maximum multiplicative depth for BP, denoted by $\tau_{\text{BP}}$, is reached when updating the first layer $h = 1$. In contrast, ReBoot partitions the training into independent encrypted local-loss blocks, and gradients are propagated only within individual blocks, as explained in Section~\ref{subsec:architecture}. Thus, the maximum multiplicative depth of ReBoot, denoted by $\tau_{\text{ReBoot}}$, is determined by the encrypted computation at the deepest layer $h = H$. 

Table~\ref{tab:depth} details the multiplicative depth for both BP and ReBoot, including both the forward and backward passes. Specifically, for a \textit{RE-Block}, the forward pass depth is $\tau^{\text{fw}} = 2$, comprising one multiplication from the \textit{RE-Matmul} of the \emph{RE-FC} layer and one from the \emph{EncryptedPolyReLU} activation function. The backward pass multiplicative depth is $\tau^{\text{bw}} = 4$, consisting of one multiplication from the \textit{RE-Matmul} of the \emph{CE-FC} local classifier, one from the derivative of the activation function \emph{EncryptedPolyReLU}$^{\, \prime}$, and two from the weight update procedure. In contrast, for a \textit{CE-Block}, the forward pass multiplicative depth is $\tau^{\text{fw}} = 3$, with two multiplications from the \textit{CE-Matmul} of the \emph{CE-FC} layer and one from the \emph{EncryptedPolyReLU} activation function. The backward pass depth is $\tau^{\text{bw}} = 5$, comprising two multiplications from the \textit{CE-Matmul} of the \emph{RE-FC} local classifier, one from the derivative of the activation function \emph{EncryptedPolyReLU}$^{\, \prime}$, and two from the weight update procedure. The cumulative multiplicative depth of ReBoot at a generic layer $h\in\mathcal{H}$ can be expressed as $\tau^{\text{fw}_h}_{\text{ReBoot}} = \lfloor 2.5 \cdot h \rfloor$ for the forward computation, and $\tau^{\text{bw}_h}_{\text{ReBoot}} = \lfloor 4.5 + 1 - (h \bmod 2) \rfloor$ for the backward computation. It is worth noting that the total multiplicative depth at layer $h$ is given by $\tau_{\text{ReBoot}}^h = \tau^{\text{fw}_h}_{\text{ReBoot}} + \tau^{\text{bw}_h}_{\text{ReBoot}}$. 

The reduction in maximum multiplicative depth achieved by ReBoot compared to BP can thus be quantified as
\begin{equation*}
    \Delta \tau = \tau_{\text{BP}} - \tau_{\text{ReBoot}} = \lfloor 2.5 \cdot H \rfloor - 3 - (H \mod 2).
\end{equation*}
This expression formalizes the depth savings and highlights ReBoot's scalability advantage. As the number of hidden layers $H$ in the NN increases, the maximum depth gap $\Delta \tau$ widens, significantly reducing ciphertext noise accumulation and frequency of bootstrapping.


\section{Experimental Results}
\label{sec:experimental_results}
In this section, we evaluate both the effectiveness and efficiency of ReBoot in comparison to existing solutions in the literature. Specifically, Section~\ref{subsec:exp_setting} outlines the experimental setup, including the datasets, NN architectures, and CKKS encryption parameters used. Section~\ref{subsec:exp_precision} examines the impact of CKKS approximate computation on the precision of the encrypted training process, while Section~\ref{subsec:exp_accuracy} evaluates ReBoot accuracy in comparison to standard MLPs trained using FP32-precision plaintext BP. Section~\ref{subsec:sota_comparison} presents a comparative analysis of ReBoot accuracy against state-of-the-art (SotA) solutions. Finally, Section~\ref{subsec:exp_demands} provides an in-depth assessment of ReBoot computational requirements, benchmarking its training latency against related encrypted DNN frameworks.

\subsection{Experimental Setup}
\label{subsec:exp_setting}
The experimental campaign is based on publicly available benchmarks, comprising $5$ image classification datasets and $5$ tabular datasets, as listed in Table~\ref{table:datasets}. All experiments were conducted on an Ubuntu 20.04 LTS workstation equipped with two Intel Xeon Gold 5318S CPUs and 384 GB of RAM. The set of hyperparameters used for each experiment are provided in Appendix~\ref{sec:appendix_experiment_hyperparams}. The code implementing ReBoot is made available as a public repository.

\begin{table}[t]
  \footnotesize
  \caption{Datasets considered in the experimental campaign.}
  \label{table:datasets}
  \centering
  \begin{tabular}{cccccc}
    \toprule
    Dataset& Type& \makecell{Input size}&Classes&\makecell{Train\\ size}&\makecell{Test\\ size}\\
    \midrule
    MNIST \citep{Deng2012Mnist}& Image& $14 \times 14$& $10$& $60,000$&$10,000$\\
    Fashion-MNIST \citep{Xiao2017Fashionmnist} & Image& $14 \times 14$& $10$& $60,000$&$10,000$\\
    Kuzushiji-MNIST \citep{Clanuwat2018Deep} & Image& $14 \times 14$& $10$& $60,000$&$10,000$\\
    C-MNIST \citep{Nandakumar2019Towards} & Image& $8 \times 8$& $10$& $60,000$&$10,000$\\
    T-MNIST \citep{Colombo2024Training} & Image& $4 \times 4$& $3$& $50$&$3,147$\\
    \midrule
    Letter Recognition \citep{Slate1991Letter} & Tabular& 16& $26$& $16,000$&$4,000$\\
    Breast Cancer \citep{Wolberg1993Breast} & Tabular& $30$& $2$& $455$&$114$\\
    Heart Disease \citep{Janosi1989Heart} & Tabular& $13$& $5$& $242$&$61$\\
    Penguins \citep{Allison2020Penguins} & Tabular& $6$& $3$& $275$&$69$\\
    Iris \citep{Fisher1936Iris} & Tabular& $4$ & $3$ & $120$&$30$\\
    \bottomrule 
  \end{tabular}
\end{table}

\begin{table}[t]
  \footnotesize
  \caption{MLP architectures and encryption parameters. $N$ denotes the polynomial degree, $\Delta$ the scaling factor, and $l = \tau_{\text{ReBoot}} + \tau_{\text{bs}}$ the scheme level, where $\tau_{\text{ReBoot}}$ and $\tau_{\text{bs}}$ represent the maximum multiplicative depth of ReBoot and the multiplicative depth of the approximate bootstrapping procedure, respectively. The dimension $k_h$ of ReBoot’s encrypted fully-connected hidden layers is indicated in parentheses.}
  \label{table:architecture}
  \centering
  \begin{tabular}{clccc}
    \toprule
    Name& \multicolumn{1}{c}{Architecture}& $N$& $l = \tau_{\text{ReBoot}} + \tau_{\text{bs}}$&$\Delta$\\
    \midrule
    \multirow{1}{*}{\emph{eMLP-1}} & \emph{RE-Block} (32) & $2^{16}$& $24 = 8 + 16$& $49$\\
    \midrule
    \emph{eMLP-2} & \emph{RE-Block} (64) $\rightarrow$ \emph{CE-Block} (32) & $2^{17}$& $27= 11 + 16$& $59$\\
    \midrule
    \multirow{2}{*}{\emph{eMLP-3}} & \emph{RE-Block} (128) $\rightarrow$ \emph{CE-Block} (64) & \multirow{2}{*}{$2^{17}$} & \multirow{2}{*}{$29= 13 + 16$} & \multirow{2}{*}{$59$}
    \\ & $\rightarrow$ \emph{RE-Block} (32) \\
    \bottomrule
  \end{tabular}
\end{table}

To assess ReBoot performance and accuracy, three NN architectures are considered, denoted as \textit{eMLP-1}, \textit{eMLP-2}, and \textit{eMLP-3}. These architectures, along with their corresponding sets of encryption parameters $\Theta$, are detailed in Table~\ref{table:architecture}. The dimensions of the \emph{RE-FC} and \emph{CE-FC} layers are indicated in parentheses. The CKKS parameters $\Theta$ are selected to ensure $\lambda = 128$ bits of security, following the recommendations of the Homomorphic Encryption Standard (HES)~\citep{Albrecht2019Homomorphic}. The scheme level $l$ is set to $l = \tau_{\text{ReBoot}} + \tau_{\text{bs}}$, where $\tau_{\text{ReBoot}}$ represents the maximum multiplicative depth required for training, as explained in Section~\ref{subsec:depth}, and $\tau_{\text{bs}}$ accounts for the additional depth needed to perform iterative bootstrapping~\citep{Bae2022META-BTS}. The polynomial degree $N$ is chosen as the smallest value that ensures 128-bit security, while the scaling factor $\Delta$ is selected to be as large as possible for the chosen $N$.

\begin{figure}[t]
    \centering
    \setlength{\abovecaptionskip}{0pt}
    \setlength{\belowcaptionskip}{5pt}
    \begin{subfigure}[htbp]{0.4\textwidth}
        \centering
        \includegraphics[width=\linewidth]{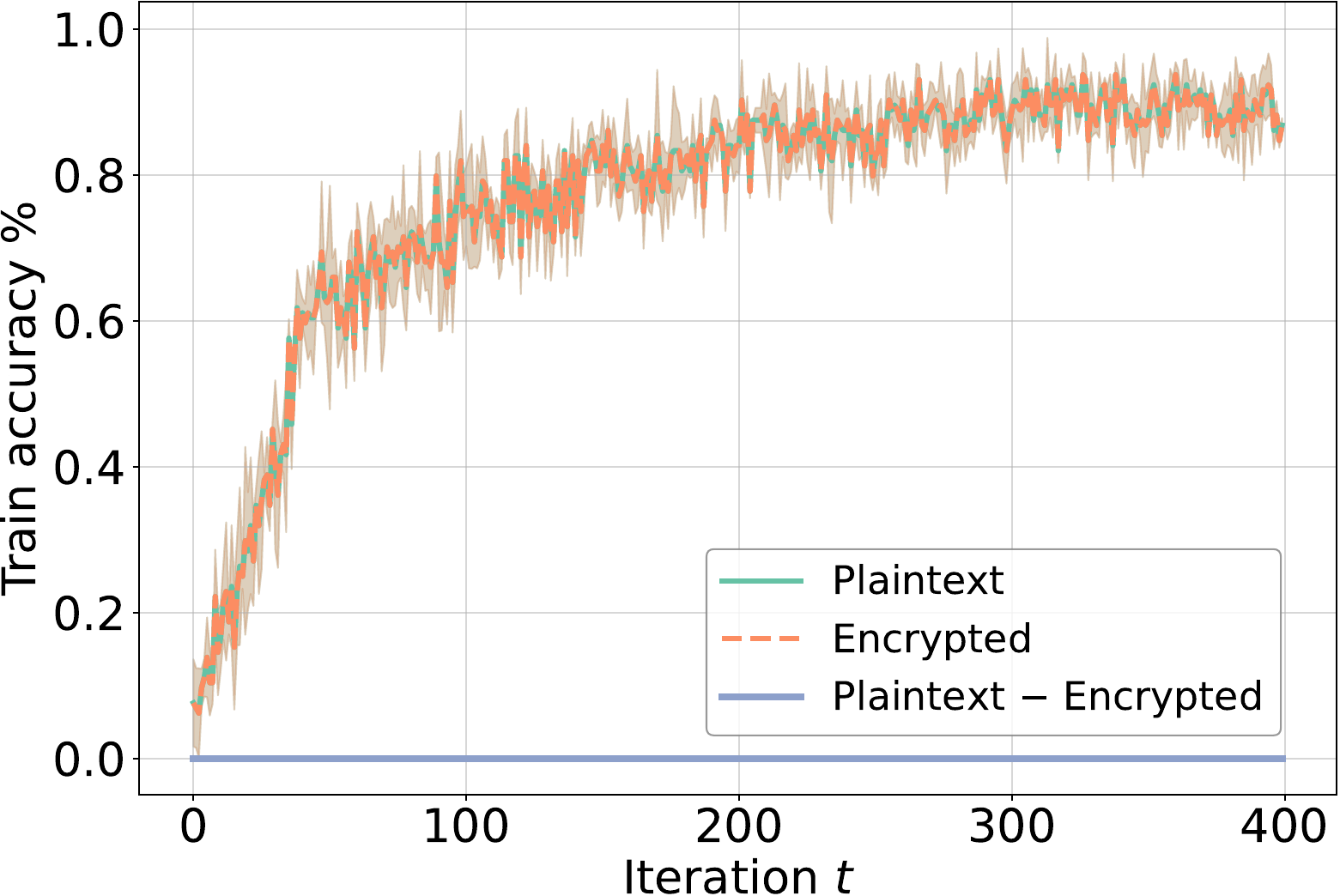}
        \caption{ReBoot encrypted and plaintext training accuracy.}
        \label{fig:precision_left}
    \end{subfigure}
    \hfill
    \begin{subfigure}[htbp]{0.4\textwidth}
        \centering
        \includegraphics[width=\linewidth]{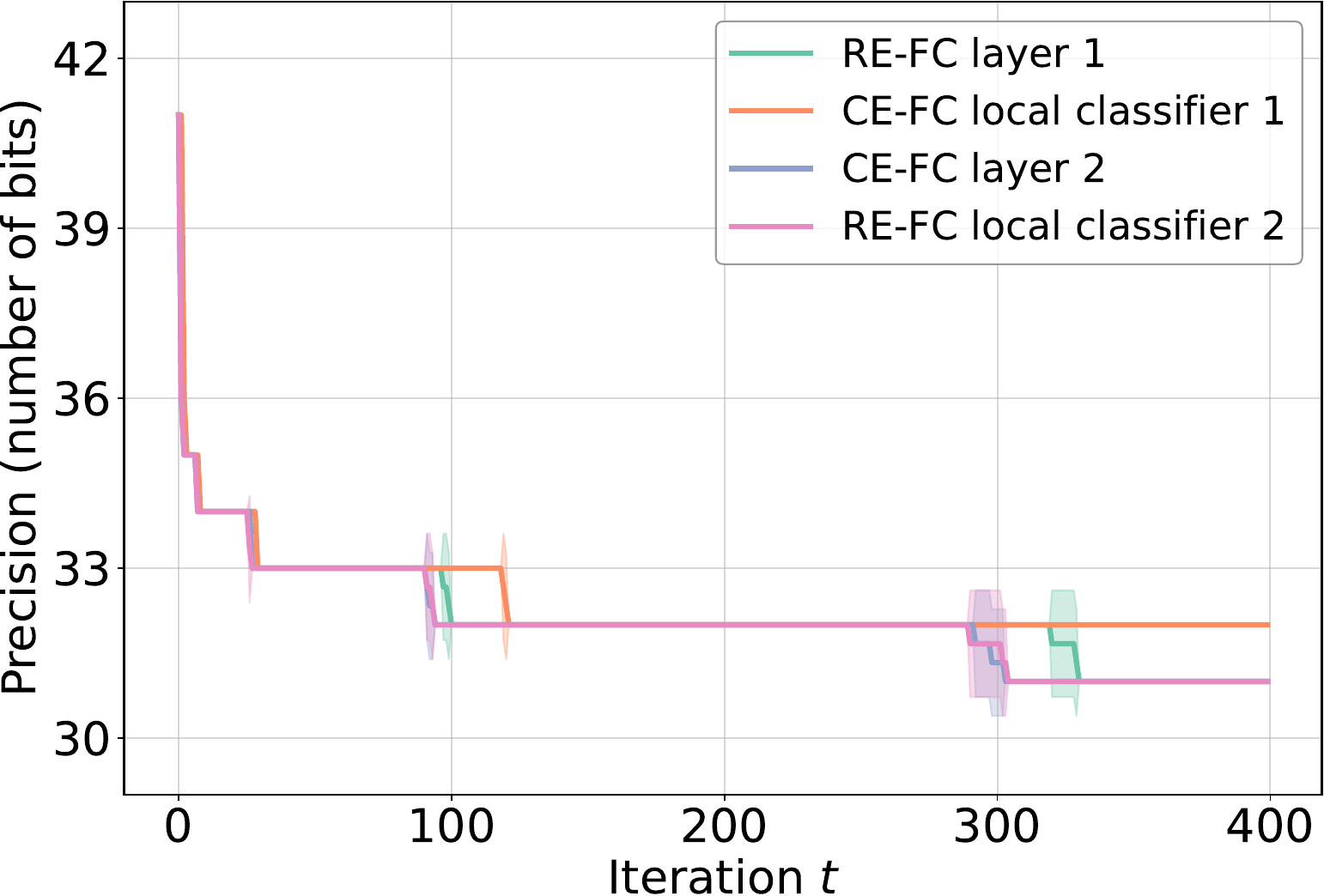}
        \caption{Weights precision per layer.}
        \label{fig:precision_right}
    \end{subfigure}
    \caption{Precision analysis of \emph{eMLP-1} on the MNIST dataset.}
    \label{fig:precision}
\end{figure}

\subsection{Precision Analysis}
\label{subsec:exp_precision}
In this section, we analyze the impact of the approximate nature of CKKS computations on ReBoot encrypted training. Each homomorphic operation in CKKS introduces noise into ciphertexts, which accumulates over time and progressively degrades their content, as discussed in Section~\ref{sec:background}. The severity of this effect depends on the scaling factor $\Delta$, which determines the initial precision of the ciphertexts, and the coefficient moduli \textbf{q}, which define how long that precision can be preserved during computation. This is particularly critical because, unlike in FHE schemes, CKKS bootstrapping does not fully refresh noise, making its careful management essential during encrypted training~\citep{Nir2022BLEACH}.

Figure~\ref{fig:precision_left} shows the training accuracy of \emph{eMLP-1} during both ReBoot plaintext and encrypted training on the MNIST dataset, averaged over three independent runs. The exact match between plaintext and encrypted training accuracy demonstrates that ReBoot encrypted training faithfully replicates the behavior of its FP32 plaintext counterpart, effectively mitigating CKKS approximation errors and managing noise accumulation. Similarly, Figure~\ref{fig:precision_right} examines the precision of CKKS ciphertexts representing the weights of \emph{eMLP-1}. Specifically, precision is measured as the number of bits in the encrypted weights that match the corresponding plaintext weights, following the method described in~\citet{Baiyu2022Securing}. The results show that precision decays logarithmically across layers, with a pronounced drop after the first bootstrapping step. This drop is primarily attributed to noise accumulation and approximation errors introduced during bootstrapping. Nevertheless, after the initial drop, the precision stabilizes in subsequent iterations, enabling ReBoot to sustain effective encrypted training.

These results confirm the functional equivalence between ReBoot encrypted and plaintext training, both in terms of learning dynamics and numerical precision. Given this demonstrated alignment, and considering the significant computational overhead of HE, subsequent ReBoot accuracy experiments are conducted using plaintext training to enable broader and more efficient evaluations across different datasets and configurations.

\begin{table*}[h]
    \small
    \caption{Test accuracy comparison across 10 independent runs between ReBoot and FP32-precision plaintext BP training.}
    \label{table:accuracy}
    \centering
    \begin{tabular}{ccccccc}
        \toprule
        \multirow{2.3}{*}{Dataset} & \multicolumn{2}{c}{\textit{eMLP-1}} & \multicolumn{2}{c}{\textit{eMLP-2}} & \multicolumn{2}{c}{\textit{eMLP-3}} \\
        \cmidrule(lr){2-3} \cmidrule(lr){4-5} \cmidrule(lr){6-7}
        & ReBoot & FP32 BP & ReBoot  & FP32 BP & ReBoot & FP32 BP \\
        \midrule
        MNIST                 & $94.61 \pm 0.27$  & $94.58\pm 0.46$   & $96.39 \pm 0.16$  & $94.18 \pm 0.32$ & $96.77 \pm 0.13$& $95.03 \pm 0.36$\\
        Fashion-MNIST         & $85.12 \pm 0.20$  & $85.09 \pm 0.22$  & $86.32 \pm 0.23$  & $82.27\pm 0.72$  & $86.64 \pm 0.28$  & $86.25 \pm 5.81$\\
        Kuzushiji-MNIST       & $76.60 \pm 0.66$  & $79.41 \pm 0.51$  & $82.31 \pm 0.53$  & $79.57\pm 0.56$  & $83.83 \pm 0.37$  & $82.07 \pm 0.51$\\
        \midrule
        Breast Cancer         & $99.03 \pm 1.78$  & $98.95 \pm 1.70$  & $98.96 \pm  1.78$  & $98.25 \pm 2.74$ & $99.03 \pm 1.88$& $98.74 \pm 2.49$\\
        Heart Disease         & $87.06 \pm 5.12$  & $87.62 \pm 5.72$  & $93.44 \pm 2.64$  & $94.87 \pm 3.00$ & $93.25 \pm 2.64$  & $94.13 \pm 2.06$\\
        Letter Recognition    & $73.08 \pm 1.27$  & $75.06 \pm 1.77$  & $85.84 \pm 0.72$  & $85.66 \pm 1.20$ & $90.66 \pm 0.90$  & $89.27 \pm 0.81$\\
        \bottomrule
    \end{tabular}
\end{table*}

\subsection{ReBoot Test Accuracy}
\label{subsec:exp_accuracy}
In this section, we evaluate the generalization performance of ReBoot in comparison to standard FP32-precision plaintext BP~\citep{Rumelhart1986LearningRB}. To this end, we train both \emph{eMLP} architectures using ReBoot and corresponding plaintext MLPs (i.e., with the same architecture) using standard BP. Specifically, plaintext models employ the NAG~\citep{Sutskever2013Importance} optimizer, the \textit{categorical cross-entropy} loss function, and the \textit{ReLU}~\citep{Agarap2019Deep} activation function, all of which are incompatible with CKKS. The complete set of hyperparameters is provided in Appendix~\ref{sec:appendix_experiment_hyperparams}. 

Table~\ref{table:accuracy} reports ReBoot's test accuracy across \emph{eMLP} architectures, compared to their corresponding standard FP32-precision plaintext BP counterparts, averaged over 10 runs. Despite operating entirely under HE constraints, ReBoot consistently achieves accuracy levels comparable to those of HE-incompatible models trained with BP. Moreover, ReBoot's test accuracy improves with increasing model complexity, indicating that its learning algorithm scales effectively with NN width and depth. These results affirm ReBoot's ability to deliver accurate and reliable encrypted training without requiring access to plaintext data or intermediate decryption, making it well-suited for privacy-preserving ML \textit{as-a-service} applications.

\subsection{Comparison with SotA Solutions}
\label{subsec:sota_comparison}
This section evaluates ReBoot's effectiveness in comparison to related work in the literature. In the context of logistic regression, several studies~\citep{Bergamaschi2019Homomorphic, Kim2018Logistic, Crockett2020Low} are limited to small-scale or proof-of-concept scenarios due to the lack of bootstrapping support. In contrast, while~\citet{Carpov2019Privacy} and~\citet{Han2018Efficient} integrate bootstrapping, they do not provide benchmarks on publicly available datasets. Given these limitations, we focus our comparison on the work of \citet{Han2018Efficient}. Results are reported in Table~\ref{table:dnn_comparison}. In particular, the MNIST dataset was reduced to a binary classification task, following the setup described in~\citet{Han2018Efficient}. ReBoot's ability to train more complex models enables it to outperform encrypted logistic regression, achieving a test accuracy improvement of $+3.27$\% using the \textit{eMLP-1} architecture. Thanks to ReBoot, encrypted training can go beyond single-layer models, enabling the training of full NNs.

In the context of DNNs, and to ensure a fair comparison with SotA solutions, ReBoot was evaluated using both the original MLP architectures from prior work and \emph{eMLP} models, which match the NN depth $H$ but feature increased layer width. Notably, ReBoot enables the use of wider fully-connected layers while preserving overall computational complexity, thereby allowing the training of more expressive models without additional cost, as analyzed in Section~\ref{subsec:exp_demands}. Table~\ref{table:dnn_comparison} summarizes the experimental results in terms of test accuracy. ReBoot consistently outperforms prior encrypted DNN methods across all evaluated datasets, with improvements of up to $+6.83\%$. In particular, while the vast majority of existing approaches rely on integer-based training, ReBoot supports real-valued training, enabling a higher-precision learning process.

\begin{table*}[htbp]
\small
  \caption{Test accuracy comparison between ReBoot and related encrypted training frameworks over 10 runs.}
  \label{table:dnn_comparison}
  \centering
  \begin{tabular}{cccccc}
    \toprule
    Dataset & Work & HE Scheme &  $\lambda$  & Architecture & Test accuracy (\%) \\
    \midrule
    \multirow{2}{*}{Binary MNIST} & \citet{Han2018Efficient} & CKKS (bootstrapping) & 128 & Logistic regression & $96.40$ \\
    & \textbf{ReBoot}  &  \textbf{CKKS (bootstrapping)} & \textbf{128} & \textbf{\emph{eMLP-1}} & \textbf{99.67} $\pm$ \textbf{0.13}\\
    \toprule
    \multirow{2}{*}{MNIST} 
      & \citet{Lou2020Glyph}        & BFV/TFHE & 80 & MLP[784-128-32-10] & 96.60 \\
      & \textbf{ReBoot}            & \textbf{CKKS (bootstrapping)} & \textbf{128} & \textbf{MLP[784-128-32-10]} & \textbf{97.67} $\pm$ \textbf{0.13} \\
    \midrule
    \multirow{3}{*}{C-MNIST}
      & \citet{Nandakumar2019Towards} & BGV & $\ll 80$ & MLP[64-32-16-10] & 96.00 \\
      & ReBoot            & CKKS (bootstrapping) & 128 & MLP[64-32-16-10] & $ 95.08 \pm 0.13 $ \\
      & \textbf{ReBoot}            & \textbf{CKKS (bootstrapping)} & \textbf{128} & \textbf{\emph{eMLP-2}} & \textbf{96.71} $\pm$ \textbf{0.12} \\
    \midrule
    \multirow{3}{*}{T-MNIST}
      & \citet{Colombo2024Training} & TFHE & 128 & MLP[16-4-2-3] & 91.90 \\
      & ReBoot         & CKKS (bootstrapping) & 128 & MLP[16-4-2-3] & $91.23 \pm 0.53$ \\
      & \textbf{ReBoot}         & \textbf{CKKS (bootstrapping)} & \textbf{128} & \textbf{\emph{eMLP-1}} & \textbf{93.39} $\pm$ \textbf{0.64} \\
    \midrule
    \multirow{2}{*}{Fashion-MNIST}
      & \citet{Colombo2024Training} & TFHE & 128 & MLP[784-200-10] & 86.00 \\
      & \textbf{ReBoot}         & \textbf{CKKS (bootstrapping)} & \textbf{128} & \textbf{MLP[784-200-10]} & \textbf{89.08} $\pm$ \textbf{0.20} \\
    \midrule
    \multirow{3}{*}{Iris}
      & \citet{Mihara2020Neural} & CKKS (leveled) & 128 & MLP[4-10-3] & 98.05 \\
      & ReBoot        & CKKS (bootstrapping) & 128 & MLP[4-10-3] & $98.44 \pm 6.40$ \\
      & \textbf{ReBoot}        & \textbf{CKKS (bootstrapping)} & \textbf{128} & \textbf{\emph{eMLP-1}} & \textbf{99.69} $\pm$ \textbf{1.88} \\
    \midrule
    \multirow{3}{*}{Penguins} 
      & \citet{Colombo2024Training} & TFHE & 128 & MLP[4-2-3] & 92.20 \\
      & ReBoot         & CKKS (bootstrapping) & 128 & MLP[4-2-3] & $81.18 \pm 2.35$ \\
      & \textbf{ReBoot}         & \textbf{CKKS (bootstrapping)} & \textbf{128} & \textbf{\emph{eMLP-1}} & \textbf{99.03} $\pm$ \textbf{0.68} \\
    \midrule
    \multirow{3}{*}{Breast Cancer}
      & \citet{Montero2024Neural} & TFHE (leveled) & 128 & MLP[30-29-1] & 98.25 \\
      & ReBoot         & CKKS (bootstrapping) & 128 & MLP[30-29-1] & $98.95 \pm 1.99$ \\
      & \textbf{ReBoot}         & \textbf{CKKS (bootstrapping)} & \textbf{128} & \textbf{\emph{eMLP-1}} & \textbf{99.03} $\pm$ \textbf{1.78} \\
    \bottomrule
  \end{tabular}
\end{table*}

\subsection{Computational and Memory Demands}
\label{subsec:exp_demands}
This section evaluates the computational and memory demands of ReBoot. Specifically, we compare ReBoot's training latency with that of related encrypted DNN frameworks. In the context of DNNs, \citet{Nandakumar2019Towards} and~\citet{Lou2020Glyph} adopt an insecure 80-bit security level, which allows the use of a smaller polynomial degree $N$, reducing computational cost but falling short of HES recommendations. Similarly, \citet{Mihara2020Neural} and \citet{Montero2024Neural} replace the bootstrapping procedure, which accounts for up to $86\%$ of the total training iteration time, with a \textit{re-encryption} mechanism. However, this mechanism requires continuous client-server interaction, making it unsuitable for \textit{as-a-service} scenarios. Given these limitations, we focus our comparison on the works of \citet{Yoo2021tBMPNet} and \citet{Colombo2024Training}, which provide a fair and meaningful basis for evaluation. 

Table~\ref{table:latency_comparison} reports the time required to execute a single training iteration $t$. Specifically, ReBoot was evaluated using both the original MLP architectures from prior work and \emph{eMLP} models. The results highlight ReBoot's significant performance advantage, achieving speedups ranging from $3.40\times$ to $8.83\times$. This gain is primarily attributed to ReBoot packing, which enables full exploitation of CKKS \textit{SIMD} operations, as explained in Section~\ref{subsec:packing}. In contrast, the lack of \textit{SIMD} support in TFHE results in sequential execution, with $\mathcal{O}(r \times c)$ complexity for a $r \times c$ weight matrix.

\begin{table*}[htbp]
\small
  \caption{Latency comparison between ReBoot and related encrypted training frameworks.}
  \label{table:latency_comparison}
  \centering
  \begin{tabular}{ccccc}
    \toprule
     Work & HE Scheme & $\lambda$ & Architecture & Training iteration time (s) \\
  \midrule
     \citet{Yoo2021tBMPNet} & TFHE & 128 & MLP[1-1-1] & 1681.20 \\
     ReBoot & CKKS (bootstrapping) & 128 & MLP[1-1-1] & 198.37 \\
     ReBoot & CKKS (bootstrapping) & 128 & \textit{eMLP-1} & 190.32 \\
     \midrule
     \citet{Colombo2024Training}& TFHE & 128 & MLP[4-2-3]& 646.72 \\
     ReBoot & CKKS (bootstrapping) & 128 & MLP[4-2-3] & 193.77 \\
     ReBoot & CKKS (bootstrapping) & 128 & \textit{eMLP-1} & 190.32 \\
     \midrule
     \citet{Colombo2024Training} & TFHE & 128 &  MLP[16-4-2-3]& 5034.00 \\
     ReBoot & CKKS (bootstrapping) & 128 & MLP[16-4-2-3] & 621.15 \\
     ReBoot & CKKS (bootstrapping) & 128 & \textit{eMLP-2} & 629.65 \\
  \bottomrule
  \end{tabular}
\end{table*}

\begin{figure}[t!]
    \centering
    \includegraphics[width=\linewidth]{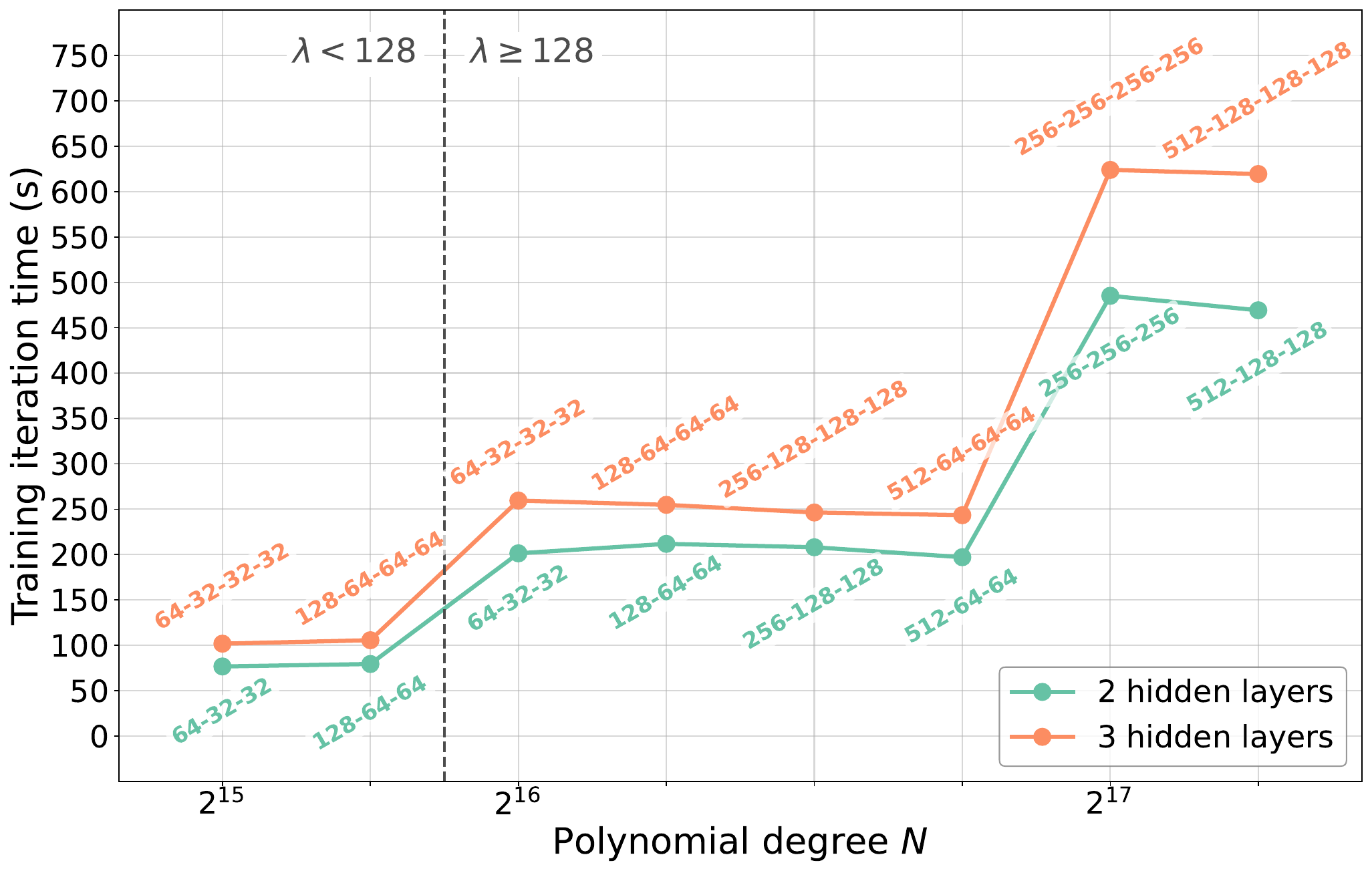}
    \caption{Computational demands of ReBoot encrypted NNs in terms of training iteration time as a function of polynomial degree $N$, layer width, and NN depth $H$.}
    \label{fig:latency}
\end{figure}

To assess the scalability of ReBoot's computational demands, we conduct experiments analyzing the impact of three key parameters on training latency: the polynomial degree $N$, the width of the NN layers $k_h$, and the overall NN depth $H$. Figure~\ref{fig:latency} shows the time required to perform a single encrypted training iteration $t$ for ReBoot architectures with $H=2$ and $H=3$ hidden layers. Experiments using $N = 2^{15}$ and $N = 2^{16}$ adopt a scaling factor $\Delta = 49$, while those with $N = 2^{17}$ use $\Delta = 59$. It is important to note that configurations with $N = 2^{15}$ do not satisfy the 128-bit security requirement and are included solely for the purpose of evaluating computational performance. Two main observations emerge. First, increasing $N$ significantly impacts latency: doubling $N$ more than doubles the  time required per training iteration. Second, for a fixed $N$, latency remains nearly constant across varying layer widths thanks to ReBoot packing, which leverages CKKS \textit{SIMD} capabilities for parallel computation. As detailed in Section~\ref{subsubsec:enc_fc}, encrypted matrix multiplications scale logarithmically with the number of neurons. However, when the layer dimension exceeds the number of available ciphertext slots, a larger $N$ is needed, resulting in a substantial increase in iteration time. In contrast, increasing the network depth has only a moderate impact, as it adds relatively few sequential encrypted operations.

Additionally, we analyze the memory usage associated with the \textit{eMLP} architectures presented in Table~\ref{table:architecture}. In general, memory consumption is primarily driven by the number of ciphertexts required to encrypt various components of the training process, including model weights, input data, intermediate activations, gradients, and output predictions. The trained encrypted models occupy 52~MB for \textit{eMLP-1}, 224~MB for \textit{eMLP-2}, and 360~MB for \textit{eMLP-3}. During training, these models reach peak memory usage of 76.07~GB, 181.45~GB, and 213.63~GB, respectively. The substantial increase in memory footprint between \textit{eMLP-1} and \textit{eMLP-2} is primarily due to the use of a larger polynomial degree $N$. Overall, memory usage scales consistently with both the NN depth and the batch size. As expected, the size of the encrypted model also grows proportionally with the number of layers: each additional layer introduces new ciphertexts not only for its weights but also for tracking momentum during optimization.

In conclusion, these experiments demonstrate that ReBoot advances the SotA in secure and non-interactive encrypted training, with computational cost largely independent of layer width (within ciphertext slot limits) and scaling moderately with NN depth. This validates ReBoot's scalability and efficiency for training deep and wide encrypted DNNs.


\section{Conclusions}
\label{sec:conclusions}
This paper introduced ReBoot, the first framework to enable fully encrypted and non-interactive training of MLPs using CKKS with bootstrapping. ReBoot proposes an advanced packing technique that significantly reduces both computational and memory overhead by leveraging CKKS \textit{SIMD} capabilities. Moreover, it introduces a dedicated HE-compliant encrypted architecture that minimizes the multiplicative depth required during training. Finally, ReBoot advances the SotA by providing a fully encrypted and non-interactive learning algorithm that enables the training of MLPs on encrypted data. Despite operating entirely under HE constraints, ReBoot achieves accuracy levels comparable to those of plaintext-trained models, demonstrating that encrypted training can be performed without compromising effectiveness.

Future work will focus on extending ReBoot to support the training of more complex DNN architectures, such as convolutional and recurrent NNs, on real-world use cases~\cite{colombo2024enhancing, tosevski2025large}. Additionally, distributed training will be explored to further reduce training latency by parallelizing the training of each encrypted local-loss block across separate machines, enabling scalable cloud-based infrastructures for privacy-preserving machine learning \textit{as-a-service} applications.


\begin{acks}
The authors would like to thank Aurora A.F. Colombo for valuable suggestions and support throughout the writing process of the ReBoot paper. This paper is supported by Dhiria s.r.l. and by PNRR-PE-AI FAIR project funded by the NextGeneration EU program.
\end{acks}

\bibliographystyle{ACM-Reference-Format}
\bibliography{main}


\begin{thebibliography}{61}


\ifx \showCODEN    \undefined \def \showCODEN     #1{\unskip}     \fi
\ifx \showDOI      \undefined \def \showDOI       #1{#1}\fi
\ifx \showISBNx    \undefined \def \showISBNx     #1{\unskip}     \fi
\ifx \showISBNxiii \undefined \def \showISBNxiii  #1{\unskip}     \fi
\ifx \showISSN     \undefined \def \showISSN      #1{\unskip}     \fi
\ifx \showLCCN     \undefined \def \showLCCN      #1{\unskip}     \fi
\ifx \shownote     \undefined \def \shownote      #1{#1}          \fi
\ifx \showarticletitle \undefined \def \showarticletitle #1{#1}   \fi
\ifx \showURL      \undefined \def \showURL       {\relax}        \fi
\providecommand\bibfield[2]{#2}
\providecommand\bibinfo[2]{#2}
\providecommand\natexlab[1]{#1}
\providecommand\showeprint[2][]{arXiv:#2}

\bibitem[Acar et~al\mbox{.}(2018)]%
        {Acar2018Survey}
\bibfield{author}{\bibinfo{person}{Abbas Acar}, \bibinfo{person}{Hidayet Aksu}, \bibinfo{person}{A.~Selcuk Uluagac}, {and} \bibinfo{person}{Mauro Conti}.} \bibinfo{year}{2018}\natexlab{}.
\newblock \showarticletitle{A Survey on Homomorphic Encryption Schemes: Theory and Implementation}.
\newblock \bibinfo{journal}{\emph{ACM Comput. Surv.}} \bibinfo{volume}{51}, \bibinfo{number}{4}, Article \bibinfo{articleno}{79} (\bibinfo{date}{jul} \bibinfo{year}{2018}), \bibinfo{numpages}{35}~pages.
\newblock
\showISSN{0360-0300}
\urldef\tempurl%
\url{https://doi.org/10.1145/3214303}
\showDOI{\tempurl}


\bibitem[Agarap(2019)]%
        {Agarap2019Deep}
\bibfield{author}{\bibinfo{person}{Abien~Fred Agarap}.} \bibinfo{year}{2019}\natexlab{}.
\newblock \bibinfo{title}{Deep Learning using Rectified Linear Units (ReLU)}.
\newblock
\newblock
\showeprint[arxiv]{1803.08375}~[cs.NE]
\urldef\tempurl%
\url{https://arxiv.org/abs/1803.08375}
\showURL{%
\tempurl}


\bibitem[Al~Badawi et~al\mbox{.}(2020)]%
        {Al2020Privft}
\bibfield{author}{\bibinfo{person}{Ahmad Al~Badawi}, \bibinfo{person}{Louie Hoang}, \bibinfo{person}{Chan~Fook Mun}, \bibinfo{person}{Kim Laine}, {and} \bibinfo{person}{Khin Mi~Mi Aung}.} \bibinfo{year}{2020}\natexlab{}.
\newblock \showarticletitle{Privft: Private and fast text classification with homomorphic encryption}.
\newblock \bibinfo{journal}{\emph{IEEE Access}}  \bibinfo{volume}{8} (\bibinfo{year}{2020}), \bibinfo{pages}{226544--226556}.
\newblock


\bibitem[Al~Badawi and Polyakov(2023)]%
        {al2023demystifying}
\bibfield{author}{\bibinfo{person}{Ahmad Al~Badawi} {and} \bibinfo{person}{Yuriy Polyakov}.} \bibinfo{year}{2023}\natexlab{}.
\newblock \showarticletitle{Demystifying bootstrapping in fully homomorphic encryption}.
\newblock \bibinfo{journal}{\emph{Cryptology ePrint Archive}} (\bibinfo{year}{2023}).
\newblock


\bibitem[Albrecht et~al\mbox{.}(2019)]%
        {Albrecht2019Homomorphic}
\bibfield{author}{\bibinfo{person}{Martin Albrecht}, \bibinfo{person}{Melissa Chase}, \bibinfo{person}{Hao Chen}, \bibinfo{person}{Jintai Ding}, \bibinfo{person}{Shafi Goldwasser}, \bibinfo{person}{Sergey Gorbunov}, \bibinfo{person}{Shai Halevi}, \bibinfo{person}{Jeffrey Hoffstein}, \bibinfo{person}{Kim Laine}, \bibinfo{person}{Kristin Lauter}, \bibinfo{person}{Satya Lokam}, \bibinfo{person}{Daniele Micciancio}, \bibinfo{person}{Dustin Moody}, \bibinfo{person}{Travis Morrison}, \bibinfo{person}{Amit Sahai}, {and} \bibinfo{person}{Vinod Vaikuntanathan}.} \bibinfo{year}{2019}\natexlab{}.
\newblock \bibinfo{title}{Homomorphic Encryption Standard}.
\newblock \bibinfo{howpublished}{Cryptology {ePrint} Archive, Paper 2019/939}.
\newblock
\urldef\tempurl%
\url{https://eprint.iacr.org/2019/939}
\showURL{%
\tempurl}


\bibitem[Ali et~al\mbox{.}(2024)]%
        {Ali2024Polynomial}
\bibfield{author}{\bibinfo{person}{Ramy~E. Ali}, \bibinfo{person}{Jinhyun So}, {and} \bibinfo{person}{A.~Salman Avestimehr}.} \bibinfo{year}{2024}\natexlab{}.
\newblock \bibinfo{title}{On Polynomial Approximations for Privacy-Preserving and Verifiable ReLU Networks}.
\newblock
\newblock
\showeprint[arxiv]{2011.05530}~[cs.LG]
\urldef\tempurl%
\url{https://arxiv.org/abs/2011.05530}
\showURL{%
\tempurl}


\bibitem[Badawi et~al\mbox{.}(2022)]%
        {Ahmad2022OpenFHE}
\bibfield{author}{\bibinfo{person}{Ahmad~Al Badawi}, \bibinfo{person}{Andreea Alexandru}, \bibinfo{person}{Jack Bates}, \bibinfo{person}{Flavio Bergamaschi}, \bibinfo{person}{David~Bruce Cousins}, \bibinfo{person}{Saroja Erabelli}, \bibinfo{person}{Nicholas Genise}, \bibinfo{person}{Shai Halevi}, \bibinfo{person}{Hamish Hunt}, \bibinfo{person}{Andrey Kim}, \bibinfo{person}{Yongwoo Lee}, \bibinfo{person}{Zeyu Liu}, \bibinfo{person}{Daniele Micciancio}, \bibinfo{person}{Carlo Pascoe}, \bibinfo{person}{Yuriy Polyakov}, \bibinfo{person}{Ian Quah}, \bibinfo{person}{Saraswathy R.V.}, \bibinfo{person}{Kurt Rohloff}, \bibinfo{person}{Jonathan Saylor}, \bibinfo{person}{Dmitriy Suponitsky}, \bibinfo{person}{Matthew Triplett}, \bibinfo{person}{Vinod Vaikuntanathan}, {and} \bibinfo{person}{Vincent Zucca}.} \bibinfo{year}{2022}\natexlab{}.
\newblock \bibinfo{title}{{OpenFHE}: Open-Source Fully Homomorphic Encryption Library}.
\newblock \bibinfo{howpublished}{Cryptology ePrint Archive, Paper 2022/915}.
\newblock
\urldef\tempurl%
\url{https://eprint.iacr.org/2022/915}
\showURL{%
\tempurl}
\newblock
\shownote{\url{https://eprint.iacr.org/2022/915}}.


\bibitem[Bae et~al\mbox{.}(2022)]%
        {Bae2022META-BTS}
\bibfield{author}{\bibinfo{person}{Youngjin Bae}, \bibinfo{person}{Jung~Hee Cheon}, \bibinfo{person}{Wonhee Cho}, \bibinfo{person}{Jaehyung Kim}, {and} \bibinfo{person}{Taekyung Kim}.} \bibinfo{year}{2022}\natexlab{}.
\newblock \showarticletitle{META-BTS: Bootstrapping Precision Beyond the Limit}. In \bibinfo{booktitle}{\emph{Proceedings of the 2022 ACM SIGSAC Conference on Computer and Communications Security}} (Los Angeles, CA, USA) \emph{(\bibinfo{series}{CCS '22})}. \bibinfo{publisher}{Association for Computing Machinery}, \bibinfo{address}{New York, NY, USA}, \bibinfo{pages}{223–234}.
\newblock
\showISBNx{9781450394505}
\urldef\tempurl%
\url{https://doi.org/10.1145/3548606.3560696}
\showDOI{\tempurl}


\bibitem[Bergamaschi et~al\mbox{.}(2019)]%
        {Bergamaschi2019Homomorphic}
\bibfield{author}{\bibinfo{person}{Flavio Bergamaschi}, \bibinfo{person}{Shai Halevi}, \bibinfo{person}{Tzipora~T Halevi}, {and} \bibinfo{person}{Hamish Hunt}.} \bibinfo{year}{2019}\natexlab{}.
\newblock \showarticletitle{Homomorphic training of 30,000 logistic regression models}. In \bibinfo{booktitle}{\emph{Applied Cryptography and Network Security: 17th International Conference, ACNS 2019, Bogota, Colombia, June 5--7, 2019, Proceedings 17}}. Springer, \bibinfo{pages}{592--611}.
\newblock


\bibitem[Bishop(1995)]%
        {Bishop1995NeuralNF}
\bibfield{author}{\bibinfo{person}{Christopher~M. Bishop}.} \bibinfo{year}{1995}\natexlab{}.
\newblock \bibinfo{title}{Neural networks for pattern recognition}.
\newblock
\newblock
\urldef\tempurl%
\url{https://api.semanticscholar.org/CorpusID:60563397}
\showURL{%
\tempurl}


\bibitem[Boemer et~al\mbox{.}(2019)]%
        {Boemer2019nGraph}
\bibfield{author}{\bibinfo{person}{Fabian Boemer}, \bibinfo{person}{Anamaria Costache}, \bibinfo{person}{Rosario Cammarota}, {and} \bibinfo{person}{Casimir Wierzynski}.} \bibinfo{year}{2019}\natexlab{}.
\newblock \showarticletitle{nGraph-HE2: A High-Throughput Framework for Neural Network Inference on Encrypted Data}. In \bibinfo{booktitle}{\emph{Proceedings of the 7th ACM Workshop on Encrypted Computing \& Applied Homomorphic Cryptography}} (London, United Kingdom) \emph{(\bibinfo{series}{WAHC'19})}. \bibinfo{publisher}{Association for Computing Machinery}, \bibinfo{address}{New York, NY, USA}, \bibinfo{pages}{45–56}.
\newblock
\showISBNx{9781450368292}
\urldef\tempurl%
\url{https://doi.org/10.1145/3338469.3358944}
\showDOI{\tempurl}


\bibitem[Bossuat et~al\mbox{.}(2021)]%
        {Bossuat2021Efficient}
\bibfield{author}{\bibinfo{person}{Jean-Philippe Bossuat}, \bibinfo{person}{Christian Mouchet}, \bibinfo{person}{Juan Troncoso-Pastoriza}, {and} \bibinfo{person}{Jean-Pierre Hubaux}.} \bibinfo{year}{2021}\natexlab{}.
\newblock \showarticletitle{Efficient Bootstrapping for Approximate Homomorphic Encryption with Non-sparse Keys}. In \bibinfo{booktitle}{\emph{Advances in Cryptology -- EUROCRYPT 2021}}, \bibfield{editor}{\bibinfo{person}{Anne Canteaut} {and} \bibinfo{person}{Fran{\c{c}}ois-Xavier Standaert}} (Eds.). \bibinfo{publisher}{Springer International Publishing}, \bibinfo{address}{Cham}, \bibinfo{pages}{587--617}.
\newblock
\showISBNx{978-3-030-77870-5}


\bibitem[Brakerski et~al\mbox{.}(2012)]%
        {Brakerski2012Leveled}
\bibfield{author}{\bibinfo{person}{Zvika Brakerski}, \bibinfo{person}{Craig Gentry}, {and} \bibinfo{person}{Vinod Vaikuntanathan}.} \bibinfo{year}{2012}\natexlab{}.
\newblock \showarticletitle{(Leveled) fully homomorphic encryption without bootstrapping}. In \bibinfo{booktitle}{\emph{Proceedings of the 3rd Innovations in Theoretical Computer Science Conference}} (Cambridge, Massachusetts) \emph{(\bibinfo{series}{ITCS '12})}. \bibinfo{publisher}{Association for Computing Machinery}, \bibinfo{address}{New York, NY, USA}, \bibinfo{pages}{309–325}.
\newblock
\showISBNx{9781450311151}
\urldef\tempurl%
\url{https://doi.org/10.1145/2090236.2090262}
\showDOI{\tempurl}


\bibitem[Brakerski et~al\mbox{.}(2013)]%
        {brakerski2013classical}
\bibfield{author}{\bibinfo{person}{Zvika Brakerski}, \bibinfo{person}{Adeline Langlois}, \bibinfo{person}{Chris Peikert}, \bibinfo{person}{Oded Regev}, {and} \bibinfo{person}{Damien Stehl{\'e}}.} \bibinfo{year}{2013}\natexlab{}.
\newblock \showarticletitle{Classical hardness of learning with errors}. In \bibinfo{booktitle}{\emph{Proceedings of the forty-fifth annual ACM symposium on Theory of computing}}. \bibinfo{pages}{575--584}.
\newblock


\bibitem[Carpov et~al\mbox{.}(2019)]%
        {Carpov2019Privacy}
\bibfield{author}{\bibinfo{person}{Sergiu Carpov}, \bibinfo{person}{Nicolas Gama}, \bibinfo{person}{Mariya Georgieva}, {and} \bibinfo{person}{Juan~Ramon Troncoso-Pastoriza}.} \bibinfo{year}{2019}\natexlab{}.
\newblock \bibinfo{title}{Privacy-preserving semi-parallel logistic regression training with Fully Homomorphic Encryption}.
\newblock \bibinfo{howpublished}{Cryptology {ePrint} Archive, Paper 2019/101}.
\newblock
\urldef\tempurl%
\url{https://eprint.iacr.org/2019/101}
\showURL{%
\tempurl}


\bibitem[Casale and Roveri(2023)]%
        {casale2023scheduling}
\bibfield{author}{\bibinfo{person}{Giuliano Casale} {and} \bibinfo{person}{Manuel Roveri}.} \bibinfo{year}{2023}\natexlab{}.
\newblock \showarticletitle{Scheduling inputs in early exit neural networks}.
\newblock \bibinfo{journal}{\emph{IEEE Trans. Comput.}} \bibinfo{volume}{73}, \bibinfo{number}{2} (\bibinfo{year}{2023}), \bibinfo{pages}{451--465}.
\newblock


\bibitem[Chase et~al\mbox{.}(2017)]%
        {chase2017security}
\bibfield{author}{\bibinfo{person}{Melissa Chase}, \bibinfo{person}{Hao Chen}, \bibinfo{person}{Jintai Ding}, \bibinfo{person}{Shafi Goldwasser}, \bibinfo{person}{Sergey Gorbunov}, \bibinfo{person}{Jeffrey Hoffstein}, \bibinfo{person}{Kristin Lauter}, \bibinfo{person}{Satya Lokam}, \bibinfo{person}{Dustin Moody}, \bibinfo{person}{Travis Morrison}, {et~al\mbox{.}}} \bibinfo{year}{2017}\natexlab{}.
\newblock \showarticletitle{Security of homomorphic encryption}.
\newblock \bibinfo{journal}{\emph{HomomorphicEncryption. org, Redmond WA, Tech. Rep}} (\bibinfo{year}{2017}).
\newblock


\bibitem[Cheon et~al\mbox{.}(2018)]%
        {Cheon2018Bootstrapping}
\bibfield{author}{\bibinfo{person}{Jung~Hee Cheon}, \bibinfo{person}{Kyoohyung Han}, \bibinfo{person}{Andrey Kim}, \bibinfo{person}{Miran Kim}, {and} \bibinfo{person}{Yongsoo Song}.} \bibinfo{year}{2018}\natexlab{}.
\newblock \showarticletitle{Bootstrapping for approximate homomorphic encryption}. In \bibinfo{booktitle}{\emph{Advances in Cryptology--EUROCRYPT 2018: 37th Annual International Conference on the Theory and Applications of Cryptographic Techniques, Tel Aviv, Israel, April 29-May 3, 2018 Proceedings, Part I 37}}. Springer, \bibinfo{pages}{360--384}.
\newblock


\bibitem[Cheon et~al\mbox{.}(2017)]%
        {Cheon2017Homomorphic}
\bibfield{author}{\bibinfo{person}{Jung~Hee Cheon}, \bibinfo{person}{Andrey Kim}, \bibinfo{person}{Miran Kim}, {and} \bibinfo{person}{Yongsoo Song}.} \bibinfo{year}{2017}\natexlab{}.
\newblock \showarticletitle{Homomorphic Encryption for Arithmetic of Approximate Numbers}. In \bibinfo{booktitle}{\emph{International Conference on the Theory and Application of Cryptology and Information Security}}.
\newblock
\urldef\tempurl%
\url{https://api.semanticscholar.org/CorpusID:3164123}
\showURL{%
\tempurl}


\bibitem[Chillotti et~al\mbox{.}(2018)]%
        {Chilotti2018TFHE}
\bibfield{author}{\bibinfo{person}{Ilaria Chillotti}, \bibinfo{person}{Nicolas Gama}, \bibinfo{person}{Mariya Georgieva}, {and} \bibinfo{person}{Malika Izabachène}.} \bibinfo{year}{2018}\natexlab{}.
\newblock \bibinfo{title}{{TFHE}: Fast Fully Homomorphic Encryption over the Torus}.
\newblock \bibinfo{howpublished}{Cryptology {ePrint} Archive, Paper 2018/421}.
\newblock
\urldef\tempurl%
\url{https://eprint.iacr.org/2018/421}
\showURL{%
\tempurl}


\bibitem[Clanuwat et~al\mbox{.}(2018)]%
        {Clanuwat2018Deep}
\bibfield{author}{\bibinfo{person}{Tarin Clanuwat}, \bibinfo{person}{Mikel Bober-Irizar}, \bibinfo{person}{Asanobu Kitamoto}, \bibinfo{person}{Alex Lamb}, \bibinfo{person}{Kazuaki Yamamoto}, {and} \bibinfo{person}{David Ha}.} \bibinfo{year}{2018}\natexlab{}.
\newblock \bibinfo{title}{Deep Learning for Classical Japanese Literature}.
\newblock
\newblock
\showeprint[arXiv]{cs.CV/1812.01718}~[cs.CV]


\bibitem[Colombo et~al\mbox{.}(2024a)]%
        {colombo2024enhancing}
\bibfield{author}{\bibinfo{person}{Aurora~AF Colombo}, \bibinfo{person}{Luca Colombo}, \bibinfo{person}{Alessandro Falcetta}, {and} \bibinfo{person}{Manuel Roveri}.} \bibinfo{year}{2024}\natexlab{a}.
\newblock \showarticletitle{Enhancing Privacy-Preserving Cancer Classification with Convolutional Neural Networks}. In \bibinfo{booktitle}{\emph{Biocomputing 2025: Proceedings of the Pacific Symposium}}. World Scientific, \bibinfo{pages}{565--579}.
\newblock


\bibitem[Colombo et~al\mbox{.}(2024b)]%
        {Colombo2024Training}
\bibfield{author}{\bibinfo{person}{Luca Colombo}, \bibinfo{person}{Alessandro Falcetta}, {and} \bibinfo{person}{Manuel Roveri}.} \bibinfo{year}{2024}\natexlab{b}.
\newblock \showarticletitle{Training Encrypted Neural Networks on Encrypted Data with Fully Homomorphic Encryption}. In \bibinfo{booktitle}{\emph{Proceedings of the 12th Workshop on Encrypted Computing \& Applied Homomorphic Cryptography}} (Salt Lake City, UT, USA) \emph{(\bibinfo{series}{WAHC '24})}. \bibinfo{publisher}{Association for Computing Machinery}, \bibinfo{address}{New York, NY, USA}, \bibinfo{pages}{64–75}.
\newblock
\showISBNx{9798400712418}
\urldef\tempurl%
\url{https://doi.org/10.1145/3689945.3694802}
\showDOI{\tempurl}


\bibitem[Crockett(2020)]%
        {Crockett2020Low}
\bibfield{author}{\bibinfo{person}{Eric Crockett}.} \bibinfo{year}{2020}\natexlab{}.
\newblock \showarticletitle{A low-depth homomorphic circuit for logistic regression model training}.
\newblock \bibinfo{journal}{\emph{Cryptology ePrint Archive}} (\bibinfo{year}{2020}).
\newblock


\bibitem[Deng(2012)]%
        {Deng2012Mnist}
\bibfield{author}{\bibinfo{person}{Li Deng}.} \bibinfo{year}{2012}\natexlab{}.
\newblock \showarticletitle{The mnist database of handwritten digit images for machine learning research}.
\newblock \bibinfo{journal}{\emph{IEEE Signal Processing Magazine}} \bibinfo{volume}{29}, \bibinfo{number}{6} (\bibinfo{year}{2012}), \bibinfo{pages}{141--142}.
\newblock


\bibitem[Drucker et~al\mbox{.}(2022)]%
        {Nir2022BLEACH}
\bibfield{author}{\bibinfo{person}{Nir Drucker}, \bibinfo{person}{Guy Moshkowich}, \bibinfo{person}{Tomer Pelleg}, {and} \bibinfo{person}{Hayim Shaul}.} \bibinfo{year}{2022}\natexlab{}.
\newblock \bibinfo{title}{{BLEACH}: Cleaning Errors in Discrete Computations over {CKKS}}.
\newblock \bibinfo{howpublished}{Cryptology {ePrint} Archive, Paper 2022/1298}.
\newblock
\urldef\tempurl%
\url{https://eprint.iacr.org/2022/1298}
\showURL{%
\tempurl}


\bibitem[Fan and Vercauteren(2012)]%
        {Fan2012Somewhat}
\bibfield{author}{\bibinfo{person}{Junfeng Fan} {and} \bibinfo{person}{Frederik Vercauteren}.} \bibinfo{year}{2012}\natexlab{}.
\newblock \bibinfo{title}{Somewhat Practical Fully Homomorphic Encryption}.
\newblock \bibinfo{howpublished}{Cryptology {ePrint} Archive, Paper 2012/144}.
\newblock
\urldef\tempurl%
\url{https://eprint.iacr.org/2012/144}
\showURL{%
\tempurl}


\bibitem[FISHER(1936)]%
        {Fisher1936Iris}
\bibfield{author}{\bibinfo{person}{R.~A. FISHER}.} \bibinfo{year}{1936}\natexlab{}.
\newblock \showarticletitle{THE USE OF MULTIPLE MEASUREMENTS IN TAXONOMIC PROBLEMS}.
\newblock \bibinfo{journal}{\emph{Annals of Eugenics}} \bibinfo{volume}{7}, \bibinfo{number}{2} (\bibinfo{year}{1936}), \bibinfo{pages}{179--188}.
\newblock
\urldef\tempurl%
\url{https://doi.org/10.1111/j.1469-1809.1936.tb02137.x}
\showDOI{\tempurl}
\showeprint{https://onlinelibrary.wiley.com/doi/pdf/10.1111/j.1469-1809.1936.tb02137.x}


\bibitem[Flynn(1966)]%
        {Flynn1966Very}
\bibfield{author}{\bibinfo{person}{M.J. Flynn}.} \bibinfo{year}{1966}\natexlab{}.
\newblock \showarticletitle{Very high-speed computing systems}.
\newblock \bibinfo{journal}{\emph{Proc. IEEE}} \bibinfo{volume}{54}, \bibinfo{number}{12} (\bibinfo{year}{1966}), \bibinfo{pages}{1901--1909}.
\newblock
\urldef\tempurl%
\url{https://doi.org/10.1109/PROC.1966.5273}
\showDOI{\tempurl}


\bibitem[Gentry(2009)]%
        {Gentry2009Fully}
\bibfield{author}{\bibinfo{person}{Craig Gentry}.} \bibinfo{year}{2009}\natexlab{}.
\newblock \showarticletitle{Fully homomorphic encryption using ideal lattices}. In \bibinfo{booktitle}{\emph{Proceedings of the Forty-First Annual ACM Symposium on Theory of Computing}} (Bethesda, MD, USA) \emph{(\bibinfo{series}{STOC '09})}. \bibinfo{publisher}{Association for Computing Machinery}, \bibinfo{address}{New York, NY, USA}, \bibinfo{pages}{169–178}.
\newblock
\showISBNx{9781605585062}
\urldef\tempurl%
\url{https://doi.org/10.1145/1536414.1536440}
\showDOI{\tempurl}


\bibitem[Goldwasser and Micali(2019)]%
        {goldwasser2019probabilistic}
\bibfield{author}{\bibinfo{person}{Shafi Goldwasser} {and} \bibinfo{person}{Silvio Micali}.} \bibinfo{year}{2019}\natexlab{}.
\newblock \showarticletitle{Probabilistic encryption \& how to play mental poker keeping secret all partial information}.
\newblock In \bibinfo{booktitle}{\emph{Providing sound foundations for cryptography: on the work of Shafi Goldwasser and Silvio Micali}}. \bibinfo{pages}{173--201}.
\newblock


\bibitem[Han et~al\mbox{.}(2018)]%
        {Han2018Efficient}
\bibfield{author}{\bibinfo{person}{Kyoohyung Han}, \bibinfo{person}{Seungwan Hong}, \bibinfo{person}{Jung~Hee Cheon}, {and} \bibinfo{person}{Daejun Park}.} \bibinfo{year}{2018}\natexlab{}.
\newblock \showarticletitle{Efficient logistic regression on large encrypted data}.
\newblock \bibinfo{journal}{\emph{Cryptology ePrint Archive}} (\bibinfo{year}{2018}).
\newblock


\bibitem[Horst et~al\mbox{.}(2020)]%
        {Allison2020Penguins}
\bibfield{author}{\bibinfo{person}{Allison~M Horst}, \bibinfo{person}{Alison~Presmanes Hill}, {and} \bibinfo{person}{Kristen~B Gorman}.} \bibinfo{year}{2020}\natexlab{}.
\newblock \bibinfo{booktitle}{\emph{allisonhorst/palmerpenguins: v0.1.0}}.
\newblock
\urldef\tempurl%
\url{https://doi.org/10.5281/zenodo.3960218}
\showDOI{\tempurl}


\bibitem[Janosi et~al\mbox{.}(1989)]%
        {Janosi1989Heart}
\bibfield{author}{\bibinfo{person}{Andras Janosi}, \bibinfo{person}{William Steinbrunn}, \bibinfo{person}{Matthias Pfisterer}, {and} \bibinfo{person}{Robert Detrano}.} \bibinfo{year}{1989}\natexlab{}.
\newblock \bibinfo{title}{{Heart Disease}}.
\newblock \bibinfo{howpublished}{UCI Machine Learning Repository}.
\newblock
\newblock
\shownote{{DOI}: https://doi.org/10.24432/C52P4X}.


\bibitem[Jin et~al\mbox{.}(2020)]%
        {Jin2020Secure}
\bibfield{author}{\bibinfo{person}{Chao Jin}, \bibinfo{person}{Mohamed Ragab}, {and} \bibinfo{person}{Khin Mi~Mi Aung}.} \bibinfo{year}{2020}\natexlab{}.
\newblock \showarticletitle{Secure transfer learning for machine fault diagnosis under different operating conditions}. In \bibinfo{booktitle}{\emph{International Conference on Provable Security}}. Springer, \bibinfo{pages}{278--297}.
\newblock


\bibitem[Kim et~al\mbox{.}(2018)]%
        {Kim2018Logistic}
\bibfield{author}{\bibinfo{person}{Andrey Kim}, \bibinfo{person}{Yongsoo Song}, \bibinfo{person}{Miran Kim}, \bibinfo{person}{Keewoo Lee}, {and} \bibinfo{person}{Jung~Hee Cheon}.} \bibinfo{year}{2018}\natexlab{}.
\newblock \showarticletitle{Logistic regression model training based on the approximate homomorphic encryption}.
\newblock \bibinfo{journal}{\emph{BMC medical genomics}}  \bibinfo{volume}{11} (\bibinfo{year}{2018}), \bibinfo{pages}{23--31}.
\newblock


\bibitem[Lee et~al\mbox{.}(2023)]%
        {Lee2023Hetal}
\bibfield{author}{\bibinfo{person}{Seewoo Lee}, \bibinfo{person}{Garam Lee}, \bibinfo{person}{Jung~Woo Kim}, \bibinfo{person}{Junbum Shin}, {and} \bibinfo{person}{Mun-Kyu Lee}.} \bibinfo{year}{2023}\natexlab{}.
\newblock \showarticletitle{HETAL: efficient privacy-preserving transfer learning with homomorphic encryption}. In \bibinfo{booktitle}{\emph{International Conference on Machine Learning}}. PMLR, \bibinfo{pages}{19010--19035}.
\newblock


\bibitem[Li et~al\mbox{.}(2022)]%
        {Baiyu2022Securing}
\bibfield{author}{\bibinfo{person}{Baiyu Li}, \bibinfo{person}{Daniele Micciancio}, \bibinfo{person}{Mark Schultz-Wu}, {and} \bibinfo{person}{Jessica Sorrell}.} \bibinfo{year}{2022}\natexlab{}.
\newblock \showarticletitle{Securing Approximate Homomorphic Encryption Using Differential Privacy}. In \bibinfo{booktitle}{\emph{Advances in Cryptology -- CRYPTO 2022}}, \bibfield{editor}{\bibinfo{person}{Yevgeniy Dodis} {and} \bibinfo{person}{Thomas Shrimpton}} (Eds.). \bibinfo{publisher}{Springer Nature Switzerland}, \bibinfo{address}{Cham}, \bibinfo{pages}{560--589}.
\newblock
\showISBNx{978-3-031-15802-5}


\bibitem[Lou et~al\mbox{.}(2020)]%
        {Lou2020Glyph}
\bibfield{author}{\bibinfo{person}{Qian Lou}, \bibinfo{person}{Bo Feng}, \bibinfo{person}{Geoffrey Charles~Fox}, {and} \bibinfo{person}{Lei Jiang}.} \bibinfo{year}{2020}\natexlab{}.
\newblock \showarticletitle{Glyph: Fast and Accurately Training Deep Neural Networks on Encrypted Data}. In \bibinfo{booktitle}{\emph{Advances in Neural Information Processing Systems}}, \bibfield{editor}{\bibinfo{person}{H.~Larochelle}, \bibinfo{person}{M.~Ranzato}, \bibinfo{person}{R.~Hadsell}, \bibinfo{person}{M.F. Balcan}, {and} \bibinfo{person}{H.~Lin}} (Eds.), Vol.~\bibinfo{volume}{33}. \bibinfo{publisher}{Curran Associates, Inc.}, \bibinfo{pages}{9193--9202}.
\newblock
\urldef\tempurl%
\url{https://proceedings.neurips.cc/paper_files/paper/2020/file/685ac8cadc1be5ac98da9556bc1c8d9e-Paper.pdf}
\showURL{%
\tempurl}


\bibitem[Lyubashevsky et~al\mbox{.}(2010)]%
        {Lyubashevsky2010OnIL}
\bibfield{author}{\bibinfo{person}{Vadim Lyubashevsky}, \bibinfo{person}{Chris Peikert}, {and} \bibinfo{person}{Oded Regev}.} \bibinfo{year}{2010}\natexlab{}.
\newblock \showarticletitle{On Ideal Lattices and Learning with Errors over Rings}.
\newblock \bibinfo{journal}{\emph{J. ACM}}  \bibinfo{volume}{60} (\bibinfo{year}{2010}), \bibinfo{pages}{43:1--43:35}.
\newblock
\urldef\tempurl%
\url{https://api.semanticscholar.org/CorpusID:1606347}
\showURL{%
\tempurl}


\bibitem[Mihara et~al\mbox{.}(2020)]%
        {Mihara2020Neural}
\bibfield{author}{\bibinfo{person}{Kentaro Mihara}, \bibinfo{person}{Ryohei Yamaguchi}, \bibinfo{person}{Miguel Mitsuishi}, {and} \bibinfo{person}{Yusuke Maruyama}.} \bibinfo{year}{2020}\natexlab{}.
\newblock \bibinfo{title}{Neural Network Training With Homomorphic Encryption}.
\newblock
\newblock
\showeprint[arxiv]{2012.13552}~[cs.CR]
\urldef\tempurl%
\url{https://arxiv.org/abs/2012.13552}
\showURL{%
\tempurl}


\bibitem[Montero et~al\mbox{.}(2024)]%
        {Montero2024Neural}
\bibfield{author}{\bibinfo{person}{Luis Montero}, \bibinfo{person}{Jordan Frery}, \bibinfo{person}{Celia Kherfallah}, \bibinfo{person}{Roman Bredehoft}, {and} \bibinfo{person}{Andrei Stoian}.} \bibinfo{year}{2024}\natexlab{}.
\newblock \showarticletitle{Neural Network Training on Encrypted Data with TFHE}.
\newblock \bibinfo{journal}{\emph{arXiv preprint arXiv:2401.16136}} (\bibinfo{year}{2024}).
\newblock


\bibitem[Mostafa et~al\mbox{.}(2018)]%
        {mostafa2018deep}
\bibfield{author}{\bibinfo{person}{Hesham Mostafa}, \bibinfo{person}{Vishwajith Ramesh}, {and} \bibinfo{person}{Gert Cauwenberghs}.} \bibinfo{year}{2018}\natexlab{}.
\newblock \showarticletitle{Deep supervised learning using local errors}.
\newblock \bibinfo{journal}{\emph{Frontiers in neuroscience}}  \bibinfo{volume}{12} (\bibinfo{year}{2018}), \bibinfo{pages}{608}.
\newblock


\bibitem[Nandakumar et~al\mbox{.}(2019)]%
        {Nandakumar2019Towards}
\bibfield{author}{\bibinfo{person}{Karthik Nandakumar}, \bibinfo{person}{Nalini Ratha}, \bibinfo{person}{Sharath Pankanti}, {and} \bibinfo{person}{Shai Halevi}.} \bibinfo{year}{2019}\natexlab{}.
\newblock \showarticletitle{Towards Deep Neural Network Training on Encrypted Data}. In \bibinfo{booktitle}{\emph{2019 IEEE/CVF Conference on Computer Vision and Pattern Recognition Workshops (CVPRW)}}. \bibinfo{pages}{40--48}.
\newblock
\urldef\tempurl%
\url{https://doi.org/10.1109/CVPRW.2019.00011}
\showDOI{\tempurl}


\bibitem[Nøkland and Eidnes(2019)]%
        {Nøkland2019Training}
\bibfield{author}{\bibinfo{person}{Arild Nøkland} {and} \bibinfo{person}{Lars~Hiller Eidnes}.} \bibinfo{year}{2019}\natexlab{}.
\newblock \bibinfo{title}{Training Neural Networks with Local Error Signals}.
\newblock
\newblock
\showeprint[arxiv]{1901.06656}~[stat.ML]


\bibitem[Ogburn et~al\mbox{.}(2013)]%
        {ogburn2013homomorphic}
\bibfield{author}{\bibinfo{person}{Monique Ogburn}, \bibinfo{person}{Claude Turner}, {and} \bibinfo{person}{Pushkar Dahal}.} \bibinfo{year}{2013}\natexlab{}.
\newblock \showarticletitle{Homomorphic encryption}.
\newblock \bibinfo{journal}{\emph{Procedia Computer Science}}  \bibinfo{volume}{20} (\bibinfo{year}{2013}), \bibinfo{pages}{502--509}.
\newblock


\bibitem[Panzade et~al\mbox{.}(2024)]%
        {Panzade2024Icantifinetune}
\bibfield{author}{\bibinfo{person}{Prajwal Panzade}, \bibinfo{person}{Daniel Takabi}, {and} \bibinfo{person}{Zhipeng Cai}.} \bibinfo{year}{2024}\natexlab{}.
\newblock \bibinfo{title}{I can't see it but I can Fine-tune it: On Encrypted Fine-tuning of Transformers using Fully Homomorphic Encryption}.
\newblock
\newblock
\showeprint[arxiv]{2402.09059}~[cs.LG]
\urldef\tempurl%
\url{https://arxiv.org/abs/2402.09059}
\showURL{%
\tempurl}


\bibitem[Patel et~al\mbox{.}(2023)]%
        {patel2023local}
\bibfield{author}{\bibinfo{person}{Adeetya Patel}, \bibinfo{person}{Michael Eickenberg}, {and} \bibinfo{person}{Eugene Belilovsky}.} \bibinfo{year}{2023}\natexlab{}.
\newblock \showarticletitle{Local learning with neuron groups}.
\newblock \bibinfo{journal}{\emph{arXiv preprint arXiv:2301.07635}} (\bibinfo{year}{2023}).
\newblock


\bibitem[Pirillo et~al\mbox{.}(2024)]%
        {Pirillo2024Nitro}
\bibfield{author}{\bibinfo{person}{Alberto Pirillo}, \bibinfo{person}{Luca Colombo}, {and} \bibinfo{person}{Manuel Roveri}.} \bibinfo{year}{2024}\natexlab{}.
\newblock \bibinfo{title}{NITRO-D: Native Integer-only Training of Deep Convolutional Neural Networks}.
\newblock
\newblock
\showeprint[arxiv]{2407.11698}~[cs.LG]
\urldef\tempurl%
\url{https://arxiv.org/abs/2407.11698}
\showURL{%
\tempurl}


\bibitem[Rumelhart et~al\mbox{.}(1986)]%
        {Rumelhart1986LearningRB}
\bibfield{author}{\bibinfo{person}{David~E. Rumelhart}, \bibinfo{person}{Geoffrey~E. Hinton}, {and} \bibinfo{person}{Ronald~J. Williams}.} \bibinfo{year}{1986}\natexlab{}.
\newblock \showarticletitle{Learning representations by back-propagating errors}.
\newblock \bibinfo{journal}{\emph{Nature}}  \bibinfo{volume}{323} (\bibinfo{year}{1986}), \bibinfo{pages}{533--536}.
\newblock
\urldef\tempurl%
\url{https://api.semanticscholar.org/CorpusID:205001834}
\showURL{%
\tempurl}


\bibitem[Scardapane et~al\mbox{.}(2020)]%
        {scardapane2020should}
\bibfield{author}{\bibinfo{person}{Simone Scardapane}, \bibinfo{person}{Michele Scarpiniti}, \bibinfo{person}{Enzo Baccarelli}, {and} \bibinfo{person}{Aurelio Uncini}.} \bibinfo{year}{2020}\natexlab{}.
\newblock \showarticletitle{Why should we add early exits to neural networks?}
\newblock \bibinfo{journal}{\emph{Cognitive Computation}} \bibinfo{volume}{12}, \bibinfo{number}{5} (\bibinfo{year}{2020}), \bibinfo{pages}{954--966}.
\newblock


\bibitem[Slate(1991)]%
        {Slate1991Letter}
\bibfield{author}{\bibinfo{person}{David Slate}.} \bibinfo{year}{1991}\natexlab{}.
\newblock \bibinfo{title}{{Letter Recognition}}.
\newblock \bibinfo{howpublished}{UCI Machine Learning Repository}.
\newblock
\newblock
\shownote{{DOI}: https://doi.org/10.24432/C5ZP40}.


\bibitem[Smart and Vercauteren(2014)]%
        {Smart2014Fully}
\bibfield{author}{\bibinfo{person}{N.~P. Smart} {and} \bibinfo{person}{F. Vercauteren}.} \bibinfo{year}{2014}\natexlab{}.
\newblock \showarticletitle{Fully homomorphic SIMD operations}.
\newblock \bibinfo{journal}{\emph{Des. Codes Cryptography}} \bibinfo{volume}{71}, \bibinfo{number}{1} (\bibinfo{date}{apr} \bibinfo{year}{2014}), \bibinfo{pages}{57–81}.
\newblock
\showISSN{0925-1022}
\urldef\tempurl%
\url{https://doi.org/10.1007/s10623-012-9720-4}
\showDOI{\tempurl}


\bibitem[Sutskever et~al\mbox{.}(2013)]%
        {Sutskever2013Importance}
\bibfield{author}{\bibinfo{person}{Ilya Sutskever}, \bibinfo{person}{James Martens}, \bibinfo{person}{George Dahl}, {and} \bibinfo{person}{Geoffrey Hinton}.} \bibinfo{year}{2013}\natexlab{}.
\newblock \showarticletitle{On the importance of initialization and momentum in deep learning}. In \bibinfo{booktitle}{\emph{Proceedings of the 30th International Conference on International Conference on Machine Learning - Volume 28}} (Atlanta, GA, USA) \emph{(\bibinfo{series}{ICML'13})}. \bibinfo{publisher}{JMLR.org}, \bibinfo{pages}{III–1139–III–1147}.
\newblock


\bibitem[Teerapittayanon et~al\mbox{.}(2016)]%
        {teerapittayanon2016branchynet}
\bibfield{author}{\bibinfo{person}{Surat Teerapittayanon}, \bibinfo{person}{Bradley McDanel}, {and} \bibinfo{person}{Hsiang-Tsung Kung}.} \bibinfo{year}{2016}\natexlab{}.
\newblock \showarticletitle{Branchynet: Fast inference via early exiting from deep neural networks}. In \bibinfo{booktitle}{\emph{2016 23rd international conference on pattern recognition (ICPR)}}. IEEE, \bibinfo{pages}{2464--2469}.
\newblock


\bibitem[Tosevski and Gulak(2025)]%
        {tosevski2025large}
\bibfield{author}{\bibinfo{person}{Vele Tosevski} {and} \bibinfo{person}{Glenn Gulak}.} \bibinfo{year}{2025}\natexlab{}.
\newblock \showarticletitle{Large-Scale Recurrent Neural Networks with Fully Homomorphic Encryption for Privacy-Enhanced Speaker Identification}. In \bibinfo{booktitle}{\emph{ICASSP 2025-2025 IEEE International Conference on Acoustics, Speech and Signal Processing (ICASSP)}}. IEEE, \bibinfo{pages}{1--5}.
\newblock


\bibitem[Walch et~al\mbox{.}(2022)]%
        {Walch2022Cryptotl}
\bibfield{author}{\bibinfo{person}{Roman Walch}, \bibinfo{person}{Samuel Sousa}, \bibinfo{person}{Lukas Helminger}, \bibinfo{person}{Stefanie Lindstaedt}, \bibinfo{person}{Christian Rechberger}, {and} \bibinfo{person}{Andreas Tr{\"u}gler}.} \bibinfo{year}{2022}\natexlab{}.
\newblock \showarticletitle{Cryptotl: Private, efficient and secure transfer learning}.
\newblock \bibinfo{journal}{\emph{arXiv preprint arXiv:2205.11935}} (\bibinfo{year}{2022}).
\newblock


\bibitem[Wolberg et~al\mbox{.}(1993)]%
        {Wolberg1993Breast}
\bibfield{author}{\bibinfo{person}{William Wolberg}, \bibinfo{person}{Olvi Mangasarian}, \bibinfo{person}{Nick Street}, {and} \bibinfo{person}{W. Street}.} \bibinfo{year}{1993}\natexlab{}.
\newblock \bibinfo{title}{{Breast Cancer Wisconsin (Diagnostic)}}.
\newblock \bibinfo{howpublished}{UCI Machine Learning Repository}.
\newblock
\newblock
\shownote{{DOI}: https://doi.org/10.24432/C5DW2B}.


\bibitem[Xiao et~al\mbox{.}(2017)]%
        {Xiao2017Fashionmnist}
\bibfield{author}{\bibinfo{person}{Han Xiao}, \bibinfo{person}{Kashif Rasul}, {and} \bibinfo{person}{Roland Vollgraf}.} \bibinfo{year}{2017}\natexlab{}.
\newblock \bibinfo{title}{Fashion-MNIST: a Novel Image Dataset for Benchmarking Machine Learning Algorithms}.
\newblock
\newblock
\showeprint[arxiv]{1708.07747}~[cs.LG]


\bibitem[Yoo and Yoon(2021)]%
        {Yoo2021tBMPNet}
\bibfield{author}{\bibinfo{person}{Joon~Soo Yoo} {and} \bibinfo{person}{Ji~Won Yoon}.} \bibinfo{year}{2021}\natexlab{}.
\newblock \showarticletitle{t-BMPNet: Trainable Bitwise Multilayer Perceptron Neural Network over Fully Homomorphic Encryption Scheme}.
\newblock \bibinfo{journal}{\emph{Security and Communication Networks}} \bibinfo{volume}{2021}, \bibinfo{number}{1} (\bibinfo{year}{2021}), \bibinfo{pages}{7621260}.
\newblock


\bibitem[Zhang et~al\mbox{.}(2022)]%
        {Zhang2022Privacy}
\bibfield{author}{\bibinfo{person}{Linlin Zhang}, \bibinfo{person}{Hideo Saito}, \bibinfo{person}{Liang Yang}, {and} \bibinfo{person}{Jiajie Wu}.} \bibinfo{year}{2022}\natexlab{}.
\newblock \showarticletitle{Privacy-preserving federated transfer learning for driver drowsiness detection}.
\newblock \bibinfo{journal}{\emph{IEEE Access}}  \bibinfo{volume}{10} (\bibinfo{year}{2022}), \bibinfo{pages}{80565--80574}.
\newblock


\end{thebibliography}

\appendix
\section{Experiments Hyperparameters}
\label{sec:appendix_experiment_hyperparams}
This section reports the hyperparameter configurations used in the experiments of Section~\ref{sec:experimental_results}. All ReBoot \emph{eMLP}s employ the components introduced in Section~\ref{sec:proposed_solution}, namely the ReBoot encrypted architecture, the ReBoot encrypted learning algorithm with a fixed momentum $\mu = 0.9$, the ReBoot encrypted loss function, and the \textit{EncryptedPolyReLU}. In contrast, MLPs trained using FP32-precision plaintext BP utilize the NAG~\citep{Sutskever2013Importance} optimizer, the \textit{categorical cross-entropy} loss function, and the \textit{ReLU}~\citep{Agarap2019Deep} activation function. Specifically, Table~\ref{table:exp3_hyper} and Table~\ref{table:exp1_hyper} summarize the hyperparameters used in Section~\ref{subsec:sota_comparison} and Section~\ref{subsec:exp_accuracy}, respectively.

\begin{table}[htbp]
  \small
  \caption{Hyperparameters of ReBoot for experiments of Table~\ref{table:accuracy}. $\gamma$ denotes the learning rate, $\eta$ the weight decay rate, and $b$ the batch size.}
  \label{table:exp3_hyper}
  \centering
  \begin{tabular}{ccccc}
    \toprule
     Dataset&Architecture& $\gamma$ &  $\eta$ & $b$ \\
     \midrule
     Binary MNIST & \textit{eMLP-1}& 0.001& 0.0& 48\\
     \toprule
     MNIST& MLP[784-128-32-10] & 0.0005 & 0.001 & 60 \\
      \midrule
     \multirow{2}{*}{C-MNIST}&MLP[64-32-16-10]& 0.0005 & 0.001 & 48 \\
     & \textit{eMLP-2}& 0.0005 & 0.001 & 60 \\
      \midrule
    \multirow{2}{*}{T-MNIST}& MLP[16-4-2-3]& 0.001& 0.001&10\\
     &\emph{eMLP-1}& 0.001& 0.0& 8\\
     \midrule
     Fashion-MNIST &MLP[784-200-100]& 0.00005 & 0.0   & 48 \\
      \midrule
    \multirow{2}{*}{Iris}& MLP[4-10-3]& 0.005 & 0.0 &8\\
     &\emph{eMLP-1} & 0.005 & 0.0 & 8\\
    \midrule
     \multirow{2}{*}{Penguins}& MLP[4-2-3]& 0.005& 0.001&16\\
     &\emph{eMLP-1}& 0.0005& 0.0& 8\\
     \midrule
 \multirow{2}{*}{Breast Cancer}& MLP[30-29-1]& 0.005& 0.0&16 \\
     &\emph{eMLP-1} & 0.001& 0.01& 16 \\
     \bottomrule
  \end{tabular}
\end{table}

\begin{table*}[htbp]
  \small
  \caption{Hyperparameters used for both ReBoot \emph{eMLP}s and standard MLPs trained with FP32-precision plaintext BP in the experiments of Table~\ref{table:accuracy}. $\gamma$ denotes the learning rate, $\eta$ the weight decay rate, and $b$ the batch size.}
  \label{table:exp1_hyper}
    \centering
    \begin{tabular}{cccccccccccc}
        \toprule
        \multirow{2.3}{*}{Dataset} & \multirow{2.3}{*}{Algorithm} & \multicolumn{3}{c}{\textit{eMLP-1}} & \multicolumn{3}{c}{\textit{eMLP-2}} & \multicolumn{3}{c}{\textit{eMLP-3}} \\
        \cmidrule(lr){3-5} \cmidrule(lr){6-8} \cmidrule(lr){9-11}
         & & $\gamma$ & $\eta$ & $b$ & $\gamma$ & $\eta$ & $b$ & $\gamma$ & $\eta$ & $b$ \\
        \midrule
        \multirow{2}{*}{MNIST} & ReBoot & 0.001 & 0.001 & 48 & 0.001 & 0.005 & 48 & 0.0005& 0.001& 48 \\
        & FP32 BP & 0.005& 0.001 & 48 & 0.001& 0.005 & 48 & 0.0005& 0.001& 48 \\
        
        \multirow{2}{*}{Fashion-MNIST} & ReBoot & 0.0005& 0.0& 48 & 0.0005& 0.0& 48 & 0.001& 0.005& 48 \\
        & FP32 BP & 0.005& 0.001 & 48 & 0.0005& 0.005 & 48 & 0.0005& 0.001& 48 \\
        
        \multirow{2}{*}{Kuzushiji-MNIST} & ReBoot & 0.0005& 0.01& 48 & 0.0005& 0.01& 48 & 0.0001& 0.01& 48 \\
        & FP32 BP & 0.005 & 0.001& 48 & 0.0005& 0.0& 48 & 0.0005& 0.0005& 48 \\
        \midrule        
        \multirow{2}{*}{Breast Cancer} & ReBoot & 0.005& 0.0& 16& 0.005& 0.0& 8& 0.005& 0.0& 8\\
        & FP32 BP & 0.005& 0.0& 16& 0.001 & 0.01& 16& 0.001& 0.01& 16\\
        
        \multirow{2}{*}{Heart Disease} & ReBoot & 0.0001& 0.01& 8& 0.01& 0.0& 8& 0.005& 0.0& 8\\
        & FP32 BP & 0.01& 0.005& 8& 0.005& 0.0& 8& 0.005& 0.0& 8\\

        Letter Recognition & ReBoot & 0.001 & 0.001 & 48 & 0.005& 0.001& 48 & 0.001& 0.0& 48 \\
        & FP32 BP & 0.005& 0.001& 48 & 0.005& 0.001& 48 & 0.001& 0.0& 48 \\
        \bottomrule
    \end{tabular}
\end{table*}

\end{document}